\documentclass[runningheads]{llncs}

\usepackage{eccv}

\usepackage{eccvabbrv}

\usepackage{graphicx}
\usepackage{booktabs}
\usepackage{soul}
\usepackage{float}
\usepackage[accsupp]{axessibility}  % Improves PDF readability for those with disabilities.

\usepackage[pagebackref,breaklinks,colorlinks,citecolor=eccvblue]{hyperref}
\usepackage{orcidlink}
\usepackage{amsmath}
\usepackage{booktabs}
\usepackage{multirow}
\usepackage{amssymb}  
\usepackage{array}   
\usepackage[utf8]{inputenc}
\usepackage[T1]{fontenc}
\usepackage{amsmath}
\usepackage{amssymb}
\usepackage{algorithm}
\usepackage{algpseudocode}
\usepackage{graphicx}
\usepackage{float}
\usepackage{booktabs}
\usepackage{colortbl}
\usepackage{xcolor}
\usepackage{enumitem} 
\definecolor{grayrow}{gray}{0.92}
\begin{document}
\newcommand{\wy}[1]{\textcolor{blue}{#1}}%red
\title{Robust 4D Driving Scene Reconstruction from Imperfect Visual Priors} 


\author{Xiaoyun Dong\inst{1} \and
Qian Xu\inst{1} \and
Yun Wang\inst{1} \and Yang Lu\inst{1} \and Jen-Ming Wu\inst{2} \and Jianping Wang\inst{1}}

\authorrunning{Dong et al.}

\institute{City University of Hong Kong \and Hon Hai Research Institute}

\maketitle

\begin{abstract}
Reconstructing 4D driving scenes in the wild (e.g., internet and AI-generated videos) is critical for diverse autonomous driving simulation. While recent Gaussian Scene Graph (GSG) methods achieve impressive visual quality, they heavily rely on precise priors, such as accurate camera poses and LiDAR depth, or manual annotations. When initialized with noisy priors estimated from in-the-wild videos, existing GSG methods suffer from optimization ambiguity (e.g., entangling camera and agent poses) and topological failures (e.g., missing objects), causing severe rendering artifacts. To enable robust in-the-wild reconstruction, we introduce Adaptive Gaussian Graph (AGG), a self-correcting 4D framework. Our Semantically-Guided Tick-Tock Strategy leverages 2D foundation features to explicitly decouple static background and camera pose updates from dynamic agent learning. Concurrently, our Adaptive Topology Evolution module actively rectifies graph structures by spawning missing agents, reassigning misclassified Gaussians, and pruning false positives. To rigorously evaluate this in-the-wild setting, we introduce Wild-30, a challenging benchmark of internet and generative videos. Extensive experiments on KITTI and Wild-30 validate that AGG consistently outperforms state-of-the-art approaches in visual fidelity and robustness under noisy priors. 
\end{abstract}

\section{Introduction}
\label{sec:intro}
Photorealistic 4D reconstruction~\cite{oasim, zhou2024hugsim, chen2021geosim, unisim} of driving scenes is increasingly critical for autonomous driving (AD) simulation, testing, and data generation. However, state-of-the-art urban reconstruction methods typically employ Gaussian Scene Graph (GSG)~\cite{Ost_2021_NSG, zhou2024drivinggaussian, chen2025omnire, yan2024streetgs} representation, which demands inputs that are expensive and difficult to obtain: high-precision poses (e.g., RTK-GNSS/INS), LiDAR-derived depth priors, and manually annotated dynamic agents. Even self-supervised approaches~\cite{chen2026pvg, peng2025desire, wei2025emd, huang2026s3gaussian} still require reliable camera poses and depth maps. In practice, these high-quality priors are primarily extracted from AD logs. Because these logs predominantly record routine driving, they inherently miss crucial but infrequent corner cases~\cite{scs1, scs2, scs3}. Consequently, this strict data dependency creates a major bottleneck, preventing the reconstruction of diverse simulation environments. 

\begin{figure}[t]
\includegraphics[width=0.98\textwidth]{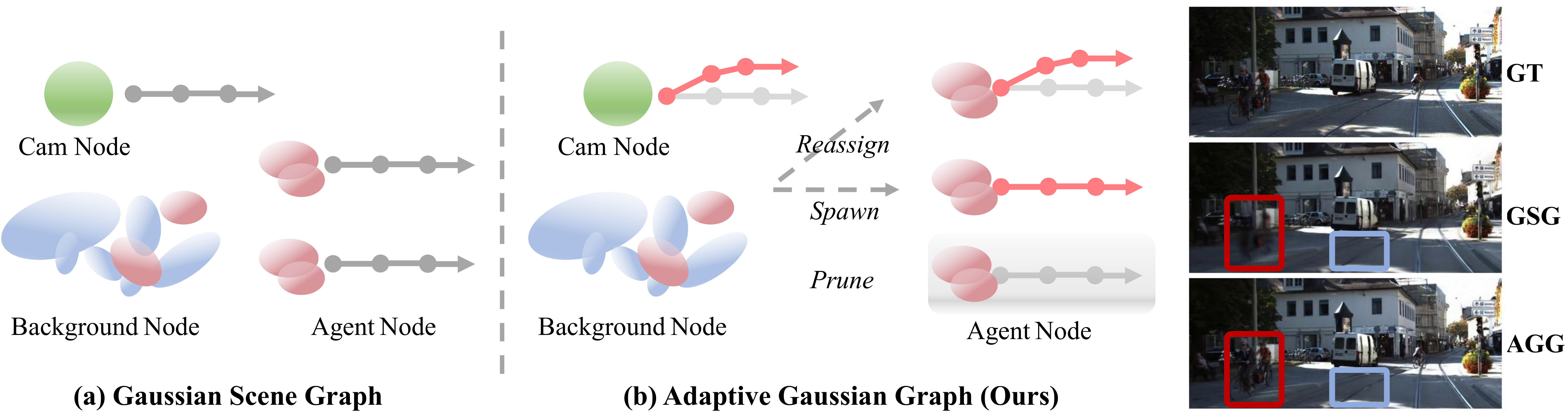}
\caption{\textbf{Reconstruction with noisy priors.} (a) GSG with inaccurate camera/agent poses and missed agents: blurred background (blue box) and missing objects (red box). (b) AGG decouples pose/background from agent learning and updates graph topology (spawn/reassign/prune), recovering agents and improving background sharpness.}
\label{fig:agg}
\end{figure}

To overcome this diversity bottleneck, we turn to pure videos in the wild—the most abundant source of driving observations (e.g., internet videos and generative model outputs). These sources are easy to collect and broaden coverage beyond routine, sensor-rich AD logs, enabling access to diverse and long-tail conditions.
Without hardware sensors and annotations, essential priors like camera poses, depth, and agent information must be inferred from off-the-shelf models~\cite{wang2025vggt, hu2025vggt4d, huang2025vipe, hu2024-DepthCrafter, must3r_cvpr25}. However, these inferred priors are inherently noisy, suffering from drifting poses, unstable depth, and undetected agents. As illustrated in Fig.~\ref{fig:agg}, when such imperfect priors are used to initialize standard GSG methods, the resulting reconstructions become brittle. Specifically, we observe two recurring failure modes: (i) optimization ambiguity, where pose errors are erroneously absorbed by background geometry or agent motions, causing severe ghosting and blurring; and (ii) topological rigidity, where dynamic agents missed during initialization can never be recovered because the graph structure is strictly fixed. Therefore, \textit{how can a GSG robustly optimize its parameters and self-correct its topology under such imperfect priors?}

To answer this question, we introduce Adaptive Gaussian Graph (AGG), a purely camera-based 4D reconstruction framework for in-the-wild driving scenes. Unlike existing methods that passively inherit errors from noisy priors, AGG proposes a novel self-correcting paradigm driven by a core principle: leveraging robust semantic priors to rectify unreliable geometric and topological priors from initialization.
We find that off-the-shelf 2D foundation models~\cite{simeoni2025dinov3} inherently provide such a semantic cue via zero-cost dense feature maps. In driving environments, these features exhibit strong semantic discriminability that enables accurate separation of static backgrounds from dynamic agents. Crucially, this dynamic-static decoupling capability generalizes consistently across highly diverse driving sequences. 

Guided by this semantic cue, AGG resolves the aforementioned failure modes through two core technical designs: First, to eliminate optimization ambiguity, we propose a Semantically-Guided Tick-Tock Strategy (Sec.~\ref{method:ticktock}). Using the decoupled semantic maps to guide gradient routing, we alternate between two disjoint optimization steps: updating camera poses and static scene geometry (Tick phase), and refining dynamic agent motions (Tock phase). This strict separation of static and dynamic optimization prevents pose drift errors from being erroneously propagated into static background or dynamic agents. Second, to overcome the limitation of fixed topological rigidity, we observe that topology errors (e.g., missed agents) trigger a distinct dual signal during training: high Gaussian gradients coupled with semantic mismatches. Driven by this signal, our proposed Adaptive Topology Evolution module (Sec.~\ref{method:topo}) enables the scene graph to dynamically spawn nodes for missing agents, reassign misclassified Gaussians, and prune false positive nodes, thus effectively self-correcting the global topology. Third, to rigorously evaluate 4D reconstruction in in-the-wild scenarios—a regime underrepresented in existing datasets—we introduce \textit{Wild-30} (Sec.~\ref{sec:wild30}). This new benchmark consists of 30 highly challenging sequences sourced from public internet driving videos~\cite{youtube} and high-fidelity generative driving videos~\cite{sora}, featuring extreme variations in geography, weather, lighting, and shaky camera motions. Our main contributions are:  
\begin{itemize}
    \item \textbf{Self-Correcting Paradigm:} We present AGG, a camera-only 4D reconstruction framework robust to noisy inputs. It pioneers a self-correcting paradigm to overcome optimization ambiguity and topological rigidity in GSG, enabling in-the-wild reconstruction for diverse driving simulation.
    \item \textbf{Semantically-Guided Tick-Tock Strategy:} We design an optimization strategy that leverages a zero-cost semantic cue to explicitly decouple camera motion and background updates from dynamic agent learning.
    \item \textbf{Adaptive Topology Evolution:} We propose a dynamic graph update mechanism driven by a dual gradient-semantic anomaly signal, enabling the graph to autonomously spawn, reassign, and prune erroneous nodes.
    \item \textbf{The Wild-30 Benchmark \& State-of-the-Art Performance:} We build Wild-30, a challenging benchmark of internet and generative videos. Experiments show that AGG achieves strong results on the KITTI dataset and remains robust on Wild-30.
\end{itemize}

\section{Related Work}
\label{sec:related_work}
\noindent\textbf{4D Scene Reconstruction for Autonomous Driving.}
Recent progress in Neural Radiance Fields (NeRF)~\cite{xie2023snerf} and 3D Gaussian Splatting (3DGS)~\cite{kerbl3Dgaussians} has significantly improved scene reconstruction for autonomous driving. Unlike early NeRFs~\cite{tancik2022blocknerf, yang2023emernerf} that suffer from slow inference, 3DGS achieves real-time rendering. Among various 3DGS approaches, the GSG representation~\cite{Ost_2021_NSG, zhou2024drivinggaussian, chen2025omnire, yan2024streetgs, zhou2024hugsim} is widely adopted for dynamic environments. By decomposing a scene into static background, dynamic agents, and ego-camera, GSG methods enable high-fidelity rendering and scene editing. However, their success heavily relies on expensive, hard-to-scale inputs: RTK camera poses, LiDAR depth, and precise 3D bounding boxes. Even self-supervised variants~\cite{chen2026pvg, peng2025desire, wei2025emd, huang2026s3gaussian} require reliable camera poses and depth priors. This dependency severely hinders scaling 4D reconstruction to unconstrained internet videos, where such precise data is unavailable.

\noindent\textbf{Video-only Priors from 4D Foundation Models.}
To break the dependency on specialized sensors and manual annotations, recent 4D foundation models infer camera motion, depth, and point tracks directly from videos~\cite{wang2025vggt, hu2025vggt4d, huang2025vipe, hu2024-DepthCrafter, must3r_cvpr25}. These predictions offer a natural initialization for camera-only reconstruction. However, compared to calibrated AD logs and manual annotations, these priors are inherently noisy, frequently suffering from drifting poses and missed agents. Directly applying them to standard GSGs causes severe ghosting and blur due to pose errors, and fails to recover agent appearances due to fixed graph structures. Therefore, the core challenge is to develop a self-correcting GSG.

To address this, we introduce the Adaptive Gaussian Graph (AGG). Using robust semantic guidance, AGG decouples the optimization of static backgrounds and dynamic agents. Furthermore, it can adaptively evolve the graph structure by spawning, reassigning, and pruning nodes. This self-correcting paradigm successfully unlocks camera-only 4D reconstruction for in-the-wild scenarios.
\begin{figure}[t]
\centering
\includegraphics[width=0.98\textwidth]{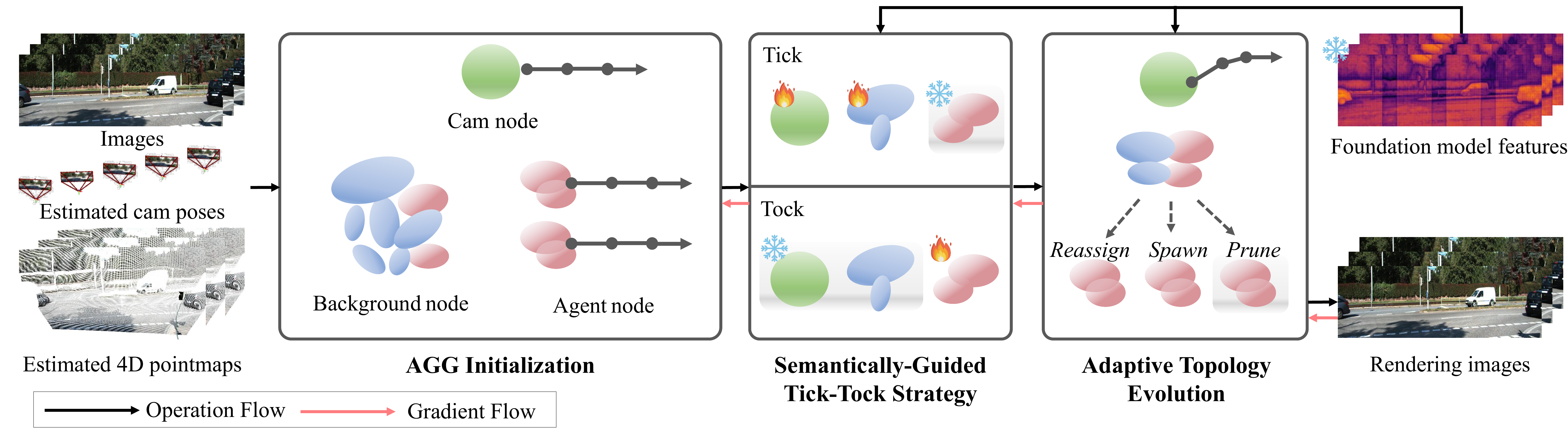}
\caption{\textbf{Overview of our AGG framework.} Initialized from noisy off-the-shelf priors, AGG optimizes the scene via a Semantically-Guided Tick-Tock Strategy that decouples the gradient flow between static (Tick) and dynamic (Tock) phases using foundation model features. Concurrently, the Adaptive Topology Evolution module continuously rectifies topology errors by explicitly spawning missing agents, reassigning mismatched Gaussians, and pruning false positives.}
\label{fig:pipeline}
\end{figure}

\section{Method}
\label{sec:method}
Our framework takes monocular or multi-view video images, along with estimated camera poses and noisy 4D pointmaps from off-the-shelf models~\cite{huang2025vipe, hu2025vggt4d, wang2025vggt, hu2024-DepthCrafter}. Calibration details are in the Appendix. As shown in Fig. \ref{fig:pipeline}, our framework proceeds in four steps. First, we initialize the AGG from noisy priors (Sec. \ref{method:agg_init}). Then, a Semantically-Guided Tick-Tock Strategy decouples static and dynamic optimization. Next, Adaptive Topology Evolution refines the graph topology by spawning, reassigning, and pruning nodes. Finally, a minimalist objective guides optimization (Sec. \ref{sec:opti}). This enables joint self-correction of geometry and topology, robustifying the reconstruction. 
 
\subsection{AGG Scene Representation}
\label{method:agg_init}

Inspired by~\cite{chen2025omnire}, AGG decomposes a dynamic scene into three components: a learnable camera node $\mathcal{N}_{cam}$, a static background node $\mathcal{N}_{bg}$, and dynamic agent nodes $\mathcal{N}_{ag}$ (categorized into rigid $\mathcal{N}_{rigid}$ and non-rigid $\mathcal{N}_{non\text{-}rigid}$ entities). The background and agent nodes are explicitly modeled using 3D Gaussians~\cite{kerbl3Dgaussians}. Each primitive is parameterized by its center $\mu_k \in \mathbb{R}^3$, covariance $\Sigma_k$ (decomposed into a scaling vector $s_k$ and rotation $q_k$), opacity $\alpha_k$, and spherical harmonics coefficients $c_k$. To establish a unified global coordinate system, we anchor the world origin to the initial camera pose $P_c(0)$. For any subsequent frame $t$ with an estimated pose $P_c(t)$, its global transformation is defined as $T(t) = P_c(0)^{-1} P_c(t)$. 

\noindent\textbf{Background Node $\mathcal{N}_{bg}$.}
We transform the static subset of the estimated 4D pointmaps at each frame $t$ into the global coordinate system via $T(t)$. The background node $\mathcal{N}_{bg}$ is then initialized from the accumulated dense point cloud.
 
\noindent\textbf{Agent Node $\mathcal{N}_{ag}(\mathcal{N}_{rigid}, \mathcal{N}_{non\text{-}rigid})$.}
We estimate agent trajectories by tracking the dense 4D pointmaps~\cite{huang2025vipe} over time. Detailed trajectory extraction and cross-frame association algorithms are deferred to the Appendix.  Denote $P_{a}(t) \in \mathrm{SE}(3)$~\cite{SE3} as the agent's estimated pose at frame $t$ in the camera coordinate system, which is derived from the estimated trajectory. The transformation matrix of the agents from the camera coordinate system to the global coordinate system is computed as $T_a(t) = T(t) P_a(t)$. 

For rigid objects $\mathcal{N}_{rigid}$ (e.g., vehicles), each node is represented by a set of 3D Gaussians $\bar{\mathcal{G}}_{rigid}$ in the canonical coordinate system. The Gaussians of objects in the global coordinate system  at time $t$ are then formulated as:
\begin{equation}
\mathcal{G}_{rigid}(t)= T_a(t) \otimes \bar{\mathcal{G}}_{rigid},
\end{equation}
where $\otimes$ performs canonical-to-global transformation on Gaussian properties. 

For non-rigid entities $\mathcal{N}_{non\text{-}rigid}$ (e.g., pedestrians), the key characteristic is their time-varying shape. To capture per-frame residuals in position, rotation, and scale, we adopt a canonical deformation model~\cite{chen2025omnire} with a learnable network $F_\phi$.  Given the object's Gaussians in canonical coordinate system $\bar{\mathcal{G}}_{non\text{-}rigid}$, their temporal evolution in the global coordinate system is:
\begin{equation}
\mathcal{G}_{non\text{-}rigid}(t) = T_a(t) \otimes (\bar{\mathcal{G}}_{non\text{-}rigid} \oplus F_\phi(t)),
\end{equation} 
where $\oplus$ integrates canonical Gaussians with predicted deformation.

\noindent\textbf{Camera Node $\mathcal{N}_{cam}$.} 
Unlike standard 3DGS~\cite{kerbl3Dgaussians}, which assumes fixed camera poses, we treat them as a sequence of learnable parameters $T_{cam}(t) \in \mathrm{SE}(3)$. These variables are directly initialized from the estimated trajectory $P_c(t)$.

\noindent\textbf{Summary of Optimizable Parameters.}
AGG jointly optimizes: background Gaussian attributes $\mathcal{N}_{bg}$ and canonical agents ($\bar{\mathcal{G}}_{rigid}$, $\bar{\mathcal{G}}_{non\text{-}rigid}$); agent poses $\mathcal{T}_a$; deformation network weights $F_\phi$; and camera poses $\mathcal{T}_{cam}$. To address optimization ambiguity and topology flaws among these coupled components, we introduce a decoupled training strategy and an adaptive topology evolution mechanism, detailed next.

\subsection{Semantically-Guided Tick-Tock Strategy}
\label{method:ticktock}

\textbf{Optimization Ambiguity Dilemma.}
Our framework jointly optimizes three coupled components ($\mathcal{N}_{cam}, \mathcal{N}_{bg},$ $\mathcal{N}_{ag}$). Directly optimizing all parameters jointly is often unstable, particularly for small or distant agents. For instance, a slight perturbation in the camera pose often forces erratic geometric updates in a distant agent to maintain photometric consistency. Such ambiguity frequently drives the optimization into suboptimal local minima, where dynamic agents become severely distorted or are entirely ignored (“ghosted”). 
%Such ambiguity typically causes the optimization to collapse into sub-optimal local minima, where dynamic agents are either heavily distorted or entirely ignored ("ghosted").

\noindent\textbf{Tick-Tock Strategy.}
To effectively decouple this joint optimization, we propose an alternating tick-tock strategy. 
During the tick phase, we freeze all dynamic agent parameters $\mathcal{N}_{ag}$. The optimizer exclusively refines $\mathcal{N}_{cam}$ and $\mathcal{N}_{bg}$, establishing a robust and stable geometric anchor between the static environment and camera poses.
Conversely, during the tock phase, we freeze $\mathcal{N}_{cam}$ and $\mathcal{N}_{bg}$, redirecting the optimization entirely to the dynamic agents. This staged optimization isolates parameter spaces, allowing us to safely employ a larger learning rate for agent poses than prior works~\cite{chen2025omnire}. Consequently, the model can efficiently correct large initialization errors inherent in noisy priors.

After an initial warm-up period of $\mathcal{E}_{warm}$ epochs, we alternate between the tick and tock phases every $\mathcal{E}_{alter}$ epochs. Upon reaching epoch $\mathcal{E}_{joint}$, we transition to a joint optimization of all variables with a decayed learning rate. Detailed hyperparameter settings are provided in Sec.~\ref{sec:exp}.

\noindent\textbf{Semantic Attention Guidance.}
To maximize optimization efficiency, we introduce a semantic attention mechanism that directs gradient flow to the relevant regions during each phase. We extract pixel-aligned features $F$ from a 2D foundation model~\cite{simeoni2025dinov3} and compress them into $F_{low} \in \mathbb{R}^{H \times W \times 64}$ via PCA~\cite{abdi2010pca}. We observe a consistent pattern across diverse driving scenes: certain PCA channels strongly activate on dynamic agents, while others reliably highlight the static environment, as illustrated in Fig.~\ref{fig:sem_att}. Consequently, we partition the PCA channels $C$ into two fixed subsets: $C_{bg}$ for the background and $C_{ag}$ for agents.

Another critical challenge is to calculate the attention maps $M_{bg}$ and $M_{ag}$ corresponding to channels subsets $C_{bg}$ for the background and $C_{ag}$ for agents from $F_{low}$, so that gradients are focused on background during the tick phase and on agents during the tock phase. A naive approach is per-channel min–max normalization, but this causes the feature scale trap: if a “dynamic” channel is inactive in the current view, local min–max still stretches this noise to $[0,1]$. During the tock phase, this falsely creates strong attention and amplifies gradients in background areas where no agents exist. 

To fix this, we use global latent normalization. For a PCA channel subset $i\in(bg,ag)$, the attention map is $  M_{i}(u) = \sum_{C_i \in C} \frac{F_{low}^{C_{i}}(u) - \min(F_{low})}{\max(F_{low}) - \min(F_{low})}$, where $\min(F_{low})$ and $\max(F_{low})$ are global extrema over all spatial positions and channels. Using a shared global scale ensures only strong activations produce high attention, suppressing noise and yielding cleaner maps without extra thresholds. 

To route gradients exclusively to the background (tick) or dynamic agents (tock), we formulate the semantic loss between the ground truth $I_{gt}$ and rendered images $I_{render}$ as:
% \begin{equation}
% \mathcal{L}_{sem} = 
% \begin{cases} 
% \sum_{u} \lambda_{sem} M_{bg}(u)  \cdot \big| I_{render}(u) - I_{gt}(u) \big|, & \text{if tick phase} \\
% \sum_{u} \lambda_{sem} M_{ag}(u) \cdot \big| I_{render}(u) - I_{gt}(u) \big|, & \text{if tock phase}
% \end{cases}
% \label{eq:lsem}
% \end{equation}
% \begin{equation}
% \mathcal{L}_{sem} =
% \begin{cases}
% \lambda_{sem}\,
% \dfrac{\sum\limits_{u} M_{bg}(u)\, \left\| I_{render}(u) - I_{gt}(u) \right\|_1}
% {\sum\limits_{u} M_{bg}(u) + \epsilon}, 
% & \text{tick phase}, \\[10pt]
% \lambda_{sem}\,
% \dfrac{\sum\limits_{u} M_{ag}(u)\, \left\| I_{render}(u) - I_{gt}(u) \right\|_1}
% {\sum\limits_{u} M_{ag}(u) + \epsilon}, 
% & \text{tock phase}.
% \end{cases}
% \label{eq:lsem}
% \end{equation}
\begin{equation}
\mathcal{L}_{sem} = \lambda_{sem}
\begin{cases} 
\frac{\sum_{u} M_{bg}(u) \| I_{render}(u) - I_{gt}(u) \|_1}{\sum_{u} M_{bg}(u) + \delta}, & \text{if tick phase} \\[12pt]
\frac{\sum_{u} M_{ag}(u) \| I_{render}(u) - I_{gt}(u) \|_1}{\sum_{u} M_{ag}(u) + \delta}, & \text{if tock phase}
\end{cases}
\label{eq:lsem}
\end{equation}
where $u$ denotes pixel coordinates, $\lambda_{sem}$ is a weighting scalar and $\delta$ is a small positive constant.
% This mechanism acts as a gradient router: amplifying background gradients during the tick phase while isolating dynamic objects during the tock phase.

\subsection{Adaptive Topology Evolution}
\label{method:topo}

\textbf{Topological Rigidity Dilemma.} GSG methods~\cite{chen2025omnire, yan2024streetgs} typically assume a fixed scene topology, relying entirely on off-the-shelf perception models for the number and initial states of agents. This rigidity makes the reconstruction highly fragile: (1) Missed agents lead to "ghosting". 
Undetected agents force static background Gaussians to compensate for their unmodeled motion, causing unnatural geometric stretching and severe blurring in affected regions (Fig.~\ref{fig:agg}-red box).
% When an agent is undetected, static background Gaussians are forced to compensate for its uncaptured motion. This causes unnatural geometric stretching and severe blurring in those regions (Fig.~\ref{fig:agg}-red box).
(2) False-positive agents produce "floating geometry". Erroneous agent nodes, initialized from misdetected bounding boxes, fit spurious Gaussians in empty space.
To break this rigidity, we introduce an adaptive topology control mechanism that dynamically spawns, reassigns, and prunes agent nodes during training. Instead of purely refining continuous variables such as poses and Gaussian attributes, AGG elevates the process to joint topology-parameter optimization, automatically correcting the number of agents and their spatial assignments.

\noindent\textbf{Prototype-based Semantic Priors.}
Unlike existing approaches~\cite{Zhou2024HUGS, zhou2024hugsim} that rely on rigid binary masks, we robustly distinguish dynamic entities from the background Gaussians by comparing the feature similarity with the semantic prototype bank $B$. Constructed from the Cityscapes dataset~\cite{Cordts2016Cityscapes}, this bank offers stronger generalization than standard closed-set classifiers. Specifically, we aggregate feature vectors for rigid agents (e.g., vehicles), non-rigid entities (e.g., pedestrians), and background regions, computing their feature centroids to serve as prototype representations, denoted as $B = \{B_{rigid}, B_{non-rigid}, B_{bg}\}$. We denote $B_{ag}= \{B_{rigid}, B_{non-rigid} \}$. For a given Gaussian $g$ with projected 2D feature $F_t(g) \in \mathbb{R}^D$ of a pre-trained foundation model~\cite{simeoni2025dinov3}, we compute its semantic score $P_v(g)$ compared to the prototype bank $B_v$ based on the cosine similarity: $P_v(g) = \mathrm{sim}(F_t(g), B_v) =  \frac{F_t(g) \cdot B_v}{\|F_t(g)\| \|B_v\|}$. Then the Gaussian $g$ belongs to the type $v=\arg\max_{v} P_v(g)$.  
Compared to closed-set classifiers (e.g., semantic segmentation networks with a fixed label space), this prototype-plus-cosine-similarity scheme is more robust to long-tailed and unseen categories. For instance, objects such as unusually shaped trucks or fallen pedestrians, which do not appear in the training data, still tend to lie close to rigid or non-rigid prototypes in feature space and are therefore classified as agents rather than being entirely treated as background.

\begin{algorithm}[t]
\caption{Adaptive Topology Evolution (ATE)}
\label{alg:topology}
\resizebox{0.95\linewidth}{!}{ 
\begin{minipage}{\linewidth} 
\begin{algorithmic}[1]
\small
\Require Graph $\mathcal{G} = \mathcal{N}_{cam} \cup \mathcal{N}_{bg} \cup \mathcal{N}_{ag}$; Prototype bank $B$; Features $\{F_t\}_{t=1}^{T}$.
\Ensure Updated scene graph $\mathcal{G}$.

\State $T_{track} \gets \varnothing$ \Comment{Initialize tracklets}
\For{$t \gets 1$ \textbf{to} $T$}
    \State $\mathcal{G}_{grad} \gets \{ g \in \mathcal{N}_{bg} \mid \|\nabla_{\mu}L(g)\|_2 > \tau_{pos} \}$ \Comment{Gaussians with large gradient}
    \State $\mathcal{G}_{cand}^{t} \gets \left\{ g \in \mathcal{G}_{grad} \mid \mathrm{sim}(F_t(g), B_{ag}) > \max \big( \mathrm{sim}(F_t(g), B_{bg}),\, \tau_{sem} \big) \right\}$ 
    \\ \Comment{Strict semantic filter}
    \State $C_t \gets \text{DBSCAN}( \{ \mu_g \mid g \in \mathcal{G}_{cand}^{t} \} )$ \Comment{Spatial clustering}
    \State $(C^{m}_t, \mathcal{G}^{m}_t), (C^{u}_t, \mathcal{G}^{u}_t) \gets \text{MatchToAgents}(C_t, \mathcal{N}_{ag})$
    \State $\mathcal{N}_{bg} \gets \text{RemoveFromBackground}(\mathcal{N}_{bg}, \mathcal{G}^{m}_t)$  \Comment{Prune background node}
    \State $\mathcal{N}_{ag} \gets \mathcal{N}_{ag} \cup \mathcal{G}^{m}_t$ \Comment{Reassign Gaussians}
    \State $T_{track} \gets \text{UpdateTracks}(T_{track}, C^{u}_t, \mathcal{G}^{u}_t)$ \Comment{Track unmatched clusters}
\EndFor

\ForAll{tracklet $k \in T_{track}$ \textbf{with} $\mathrm{age}(k) \ge \tau_{life}$}
    \State $\mathcal{N}_{new} \gets \text{SpawnAgent}(k.\mathcal{G}, k.{poses}, k.{class})$ \Comment{Spawn missing agent}
    \State $\mathcal{N}_{ag} \gets \mathcal{N}_{ag} \cup \mathcal{N}_{new} $ \Comment{Add to scene graph}
\EndFor
\State \Return $\mathcal{G}$
\end{algorithmic}
\end{minipage}
}
\end{algorithm}
\noindent\textbf{Gradient-Induced Topology Refinement.}
We then refine the topology using both gradient magnitude and the aforementioned prototype-based semantic priors. As discussed in the previous failure mode, when an agent is missing, the background Gaussians in that region accumulate high positional gradients ($\|\nabla_{\mu}L(g)\|_2$) as they struggle to compensate for the unmodeled motion. We leverage this gradient as a robust cue to refine new agents.

As outlined in Algorithm~\ref{alg:topology}, we first extract a set of candidate primitives $\mathcal{G}_{cand}^{t}$ whose positional gradients ($\|\nabla_{\mu}L(g)\|_2 > \tau_{pos}$) , and which also pass a semantic filter: $v \in \{\text{rigid, non-rigid}\} \text{ and } \max P_v > \tau_{sem}$  where $\tau_{sem}$ is a base confidence threshold. 
Since $\mathcal{G}_{cand}^{t}$ can be spatially noisy, we group them into instance-level clusters $\mathcal{C}_t$ using DBSCAN in 3D space. We then match these clusters against the bounding boxes of existing agent nodes $\mathcal{N}_{ag}$. If a cluster falls within an existing agent's local frame, it indicates a background-agent entanglement where the agent's geometry was erroneously absorbed by the background. We correct this by explicitly pruning these matched Gaussians from $\mathcal{N}_{bg}$ and reassigning them into the corresponding $\mathcal{N}_{ag}$.

Conversely, spatially isolated clusters are treated as potential missing agents. To prevent spawning spurious nodes from transient noise, we associate and track these unmatched clusters via a Kalman Filter~\cite{kf1960}. A new agent node is spawned only if a tracklet persists for $\tau_{life}$ consecutive frames. We initialize the new node using the tracklet's smoothed trajectory and semantic class, with subsequent refinement in the tick-tock optimization (Sec.~\ref{method:ticktock}). Finally, we use standard opacity-based culling~\cite{kerbl3Dgaussians} to handle false positives: if an agent is erroneously spawned, its opacity decays to zero, and the empty node is pruned from AGG.

% Conversely, spatially isolated clusters are treated as potential missing agents. To prevent spawning spurious nodes from transient noise, we associate and track these unmatched clusters over time using a Kalman Filter~\cite{kf1960}. A new agent node is spawned only if a tracklet persists for $\tau_{life}$ consecutive frames. We initialize the new node using the tracklet's smoothed trajectory and semantic class, which is subsequently refined during the tick-tock optimization (Sec.~\ref{method:ticktock}). Finally, we rely on standard opacity-based culling~\cite{kerbl3Dgaussians} to naturally handle false positives: if an agent is erroneously spawned, its opacity decays to zero, and the empty node is pruned from the AGG.

\subsection{Optimization}
\label{sec:opti}
We supervise AGG using only the standard photometric loss $\mathcal{L}_{photo}$ and a semantic guidance loss $\mathcal{L}_{sem}$ (Eq.~\ref{eq:lsem}). The overall loss function is formulated as: $\mathcal{L} = \mathcal{L}_{photo} + \mathcal{L}_{sem}$. This minimalist objective facilitates 4D reconstruction in-the-wild scenarios. Triggering mechanisms for tick-tock strategy and topology evolution, along with specific parameter settings, are detailed in Sec.~\ref{exp:setting}.

\section{Experiments}
\label{sec:exp}
\subsection{The Wild-30 Benchmark}
\label{sec:wild30}

\begin{figure}[t]
    \centering
    \captionsetup{skip=2pt} 
    \begin{subfigure}[b]{0.49\linewidth}
        \centering
        \includegraphics[width=\linewidth]{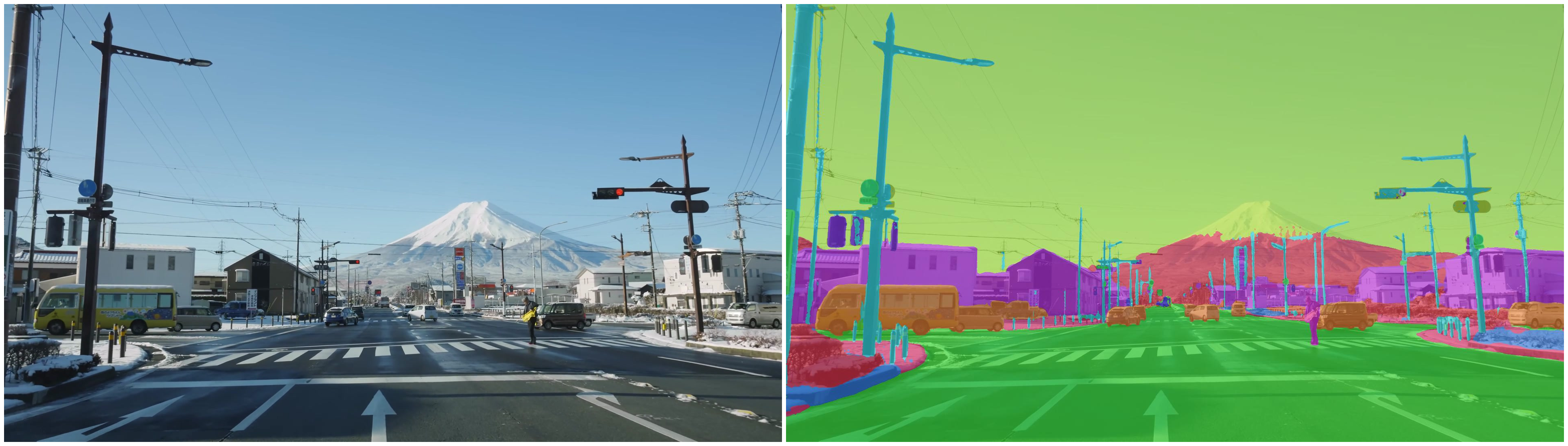} 
        \caption{Japan, sunny}
    \end{subfigure}
    \hfill 
    \begin{subfigure}[b]{0.49\linewidth}
        \centering
        \includegraphics[width=\linewidth]{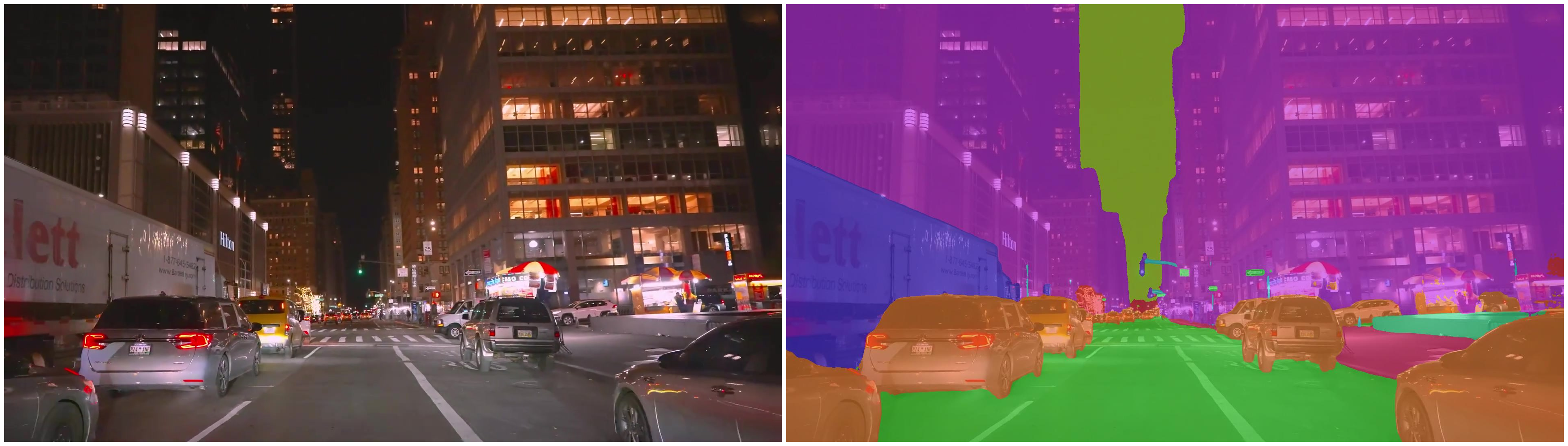} 
        \caption{America, nighttime}
    \end{subfigure}
    \begin{subfigure}[b]{0.49\linewidth}
        \centering
        \includegraphics[width=\linewidth]{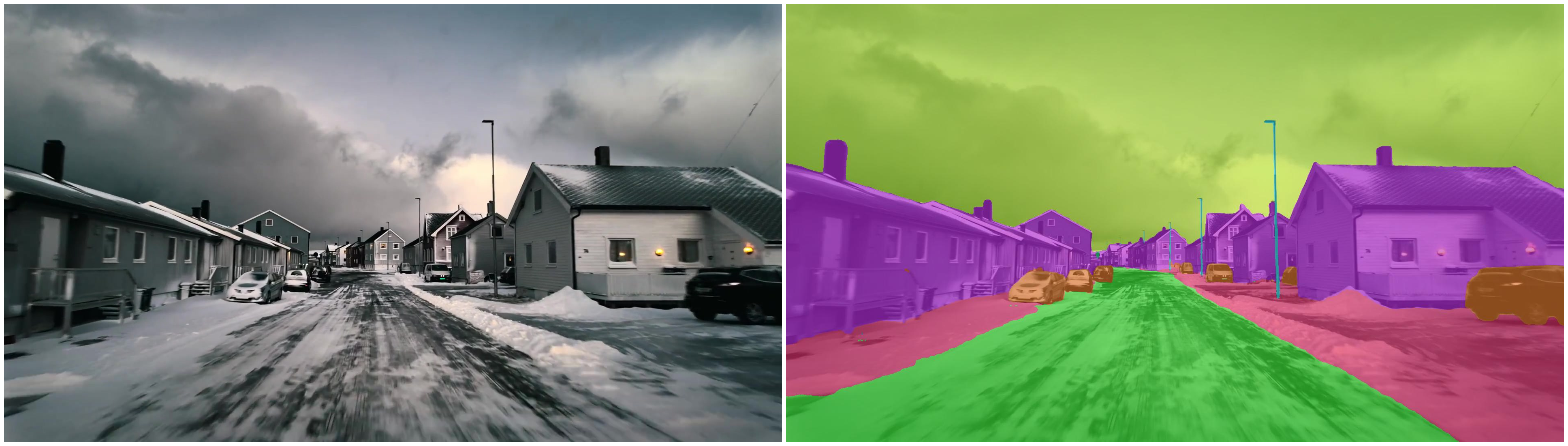}
        \caption{Norway, snowy}
        \label{fig:abl-topo}
    \end{subfigure}
    \hfill
    \begin{subfigure}[b]{0.48\linewidth}
        \centering
        \includegraphics[width=\linewidth]{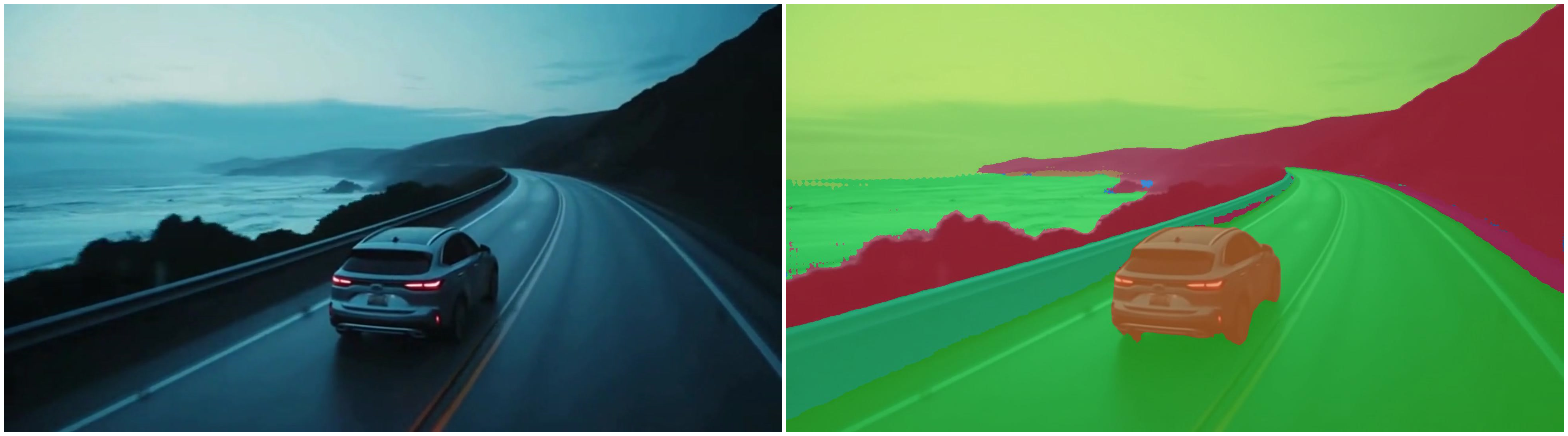}
        \caption{Generative video}
    \end{subfigure}
    \caption{\textbf{The Wild-30 benchmark.} Wild-30 features internet videos and generative videos with extreme weather (e.g., snowy) and diverse geographic locations. Auto-annotated semantic masks (right) enable agents evaluation.}
\label{fig:wild30}
\end{figure}

Most autonomous driving benchmarks~\cite{kitti, waymo} are captured with professional sensor rigs on fixed routes in a few cities. As a result, videos tend to have smooth camera motion and limited environmental diversity, and they rarely include extreme conditions such as heavy snow.  In contrast, 4D reconstruction for simulation must handle unconstrained, highly dynamic inputs, such as shaky internet videos and AI-generated videos.

To fill this gap, we introduce Wild-30, a benchmark designed to evaluate robustness under diverse scenes and unstable camera poses. Wild-30 includes 30 challenging monocular sequences: 24 real-world internet videos~\cite{youtube} (4K, 60 FPS) and 6 synthetic videos generated by state-of-the-art models~\cite{sora} (1080p, 30 FPS). The real-world split spans multiple regions (USA, Japan, China, Norway) and diverse lighting and weather conditions (sunny, rainy, snowy), while the generative split simulates dashcam footage, jointly making it a strong testbed for in-the-wild generalization. Since these videos do not provide manual 3D annotations, we use SegFormer~\cite{kerssies2025eomt} to automatically generate semantic masks for 16 common driving categories~\cite{Cordts2016Cityscapes}. The masks support fine-grained evaluation of dynamic agents, complementing standard image-level metrics. For privacy, we anonymize sensitive information such as license plates and GPS coordinates. We will release Wild-30 upon publication.

\subsection{Experimental Setups}
\label{exp:setting}
\textbf{Datasets.}
We conduct evaluations on the KITTI dataset~\cite{kitti} and our proposed Wild-30 benchmark. For KITTI, we use the highly dynamic subset from~\cite{chen2025omnire, peng2025desire, wei2025emd} and utilize both left and right cameras. To assess scalability in unconstrained environments, we use Wild-30 under a monocular setting. 

\noindent\textbf{Evaluation Protocol and Metrics.}
We evaluate scene reconstruction and novel view synthesis (NVS), holding out every 5-th frame for NVS testing. We measure image quality using standard metrics: PSNR, SSIM\cite{ssim}, and LPIPS\cite{lpips}. To explicitly assess dynamic performance, we report these metrics for full images and specifically within agent regions (defined by 3D boxes for KITTI and SegFormer masks for Wild-30). Efficiency is evaluated via rendering FPS.

\noindent\textbf{Baselines.}
We benchmark AGG against state-of-the-art methods, categorized by their required inputs: (1) Fully-supervised methods that heavily depend on accurate camera poses, LiDAR depth, and manual 3D bounding boxes (S-NeRF~\cite{xie2023snerf}, NSG~\cite{Ost_2021_NSG}, StreetGS~\cite{wei2025emd}, OmniRe~\cite{chen2025omnire}, and Mars~\cite{mars}); (2) Sensor-assisted methods that drop human annotations but still require accurate camera poses and LiDAR point clouds (SUDS~\cite{turki2023suds}, EmerNeRF~\cite{yang2023emernerf}, DeformGS~\cite{deformgs}, PVG~\cite{chen2026pvg}, DeSiRe-GS~\cite{peng2025desire}, and EMD~\cite{wei2025emd}); and (3) Pose-only methods utilizing solely exact camera poses (3DGS~\cite{kerbl3Dgaussians} and StreetSurf~\cite{guo2023streetsurf}). 
To evaluate robustness, we test all methods under an ideal setting with exact ground truth priors and a noisy setting using estimated inputs.

\noindent\textbf{Implementation Details.}
We implement AGG upon the OmniRe\cite{chen2025omnire} codebase. We extract off-the-shelf 2D features\cite{simeoni2025dinov3}, reducing them to $D=64$ dimensions via PCA. Optimization spans 30,000 iterations on a single NVIDIA A100 (80 GB) GPU, taking approximately 1 hour.

For Tick-Tock optimization, while general Gaussian attributes follow default schedules, we introduce specific adjustments for our Tick-Tock strategy. This alternating phase begins at iteration 5,000, switches every 1,000 iterations, and concludes at iteration 20,000. Benefiting from this decoupled optimization, we safely scale up the initial learning rates for agent nodes: from $1.6e^{-4}$ to $1.6e^{-3}$ for Gaussian positions ($\mu$), and from $1e^{-5}$ to $1e^{-4}$ for pose translations. Camera poses are jointly optimized with a learning rate of $1e^{-5}$. For the held-out NVS test frames, both camera and agent poses are linearly interpolated from adjacent optimized frames. The semantic loss weight is set to $\lambda_{sem}=0.8$ and $\delta = 1e^{-6}$.

We trigger this topology refinement every 10,000 iterations, terminating at iteration 20,000. During the process, we identify potential missing agents via a position-gradient threshold of $\tau_{pos}=5e^{-3}$ and a semantic similarity threshold of $\tau_{sem}=0.5$. High-gradient candidates are clustered in 3D space
% The identified high-gradient candidates are then clustered in 3D space
via DBSCAN with $\epsilon=2.5$ and $\mathrm{min\_samples}=25$. Finally, to prevent noise, we require a newly spawned agent tracklet to maintain a minimum lifespan of $\tau_{life} = 3$ frames. Detailed discussions of these parameter settings are provided in the Appendix.

\subsection{Comparisons with the State-of-the-art}
We evaluate our method against state-of-the-art baselines on the KITTI dataset and our proposed Wild-30 benchmark. Across all evaluations, AGG consistently demonstrates highly competitive reconstruction fidelity under ideal settings and exceptional robustness against noisy priors, unlocking \emph{scalable 4D reconstruction for in-the-wild scenarios}.

\begin{table}[!t]
\centering
\caption{Results on the KITTI Dataset for scene reconstruction and novel view synthesis. \textbf{Bold} indicates the best performance, and \ul{underline} indicates the second best. Note that Setting A utilizes expensive GT inputs, and Setting B uses estimated inputs.}
\label{tab:kitti-metric}
\setlength{\tabcolsep}{4.2pt} 
\renewcommand{\arraystretch}{0.8} 
\resizebox{0.85\textwidth}{!}{
\begin{tabular}{l ccc ccc c}
\toprule
\multirow{2}{*}{Method} & \multicolumn{3}{c}{Scene Reconstruction} & \multicolumn{3}{c}{Novel View Synthesis (NVS)} & \multirow{2}{*}{FPS} \\ 
\cmidrule(lr){2-4} \cmidrule(lr){5-7}
 & PSNR $\uparrow$ & SSIM $\uparrow$ & LPIPS $\downarrow$ & PSNR $\uparrow$ & SSIM $\uparrow$ & LPIPS $\downarrow$ &  \\ 
\midrule
\multicolumn{8}{l}{\textit{\textbf{Setting A:} Ideal Input (GT)}} \\ 
\midrule
NSG~\cite{Ost_2021_NSG}        & 19.19 & 0.683 & 0.189 & 17.78 & 0.645 & 0.312 & 0.19 \\
3DGS~\cite{kerbl3Dgaussians}& 21.01 & 0.811 & 0.202 & 19.54 & 0.776 & 0.224 & \textbf{125} \\
StreetSurf~\cite{guo2023streetsurf} & 24.14 & 0.819 & 0.257 & 22.48 & 0.763 & 0.304 & 0.37 \\
EmerNeRF~\cite{yang2023emernerf}   & 26.95 & 0.828 & 0.218 & 25.24 & 0.801 & 0.237 & 0.28 \\
PVG~\cite{chen2026pvg}         & 32.83 & 0.937 & 0.070 & 27.43 & 0.896 & 0.114 & \ul{59} \\
DesireGS~\cite{peng2025desire}   & 33.94 & 0.949 & \ul{0.040} & 28.87 & 0.901 & 0.106 & 41 \\
EMD~\cite{wei2025emd}        & \ul{34.13} & \ul{0.954} & \ul{0.040} & \textbf{29.05} & \ul{0.904} & \ul{0.094} & 32 \\
\rowcolor{grayrow} 
\textbf{AGG (Ours)} & \textbf{34.60} & \textbf{0.980} & \textbf{0.020} & \ul{28.99} & \textbf{0.935} & \textbf{0.037} & 35 \\ 
\midrule
% --- Setting B ---
\multicolumn{8}{l}{\textit{\textbf{Setting B:} Noisy Input}} \\ 
\midrule
PVG~\cite{chen2026pvg}    & \ul{27.14} & \ul{0.881} & 0.150 & \ul{23.32} & 0.790 & 0.175 & \ul{59} \\
DeformGS~\cite{deformgs}        & 24.13 & 0.798 & 0.207 & 22.24 & 0.746 & 0.212 &  7 \\
StreetGS~\cite{yan2024streetgs}        & 24.22 & 0.880 & \ul{0.102} & 22.13 & 0.798 & 0.115 & 33 \\
OmniRe~\cite{chen2025omnire}          & 24.07 & 0.880 & 0.111 & 22.18 & \ul{0.800} & \ul{0.114} & 31 \\
\rowcolor{grayrow} 
\textbf{AGG (Ours)}  & \textbf{29.08} & \textbf{0.943} & \textbf{0.040} & \textbf{25.25} & \textbf{0.881} & \textbf{0.062} & 35 \\ 
\bottomrule
\end{tabular}
}
\includegraphics[width=0.98\textwidth]{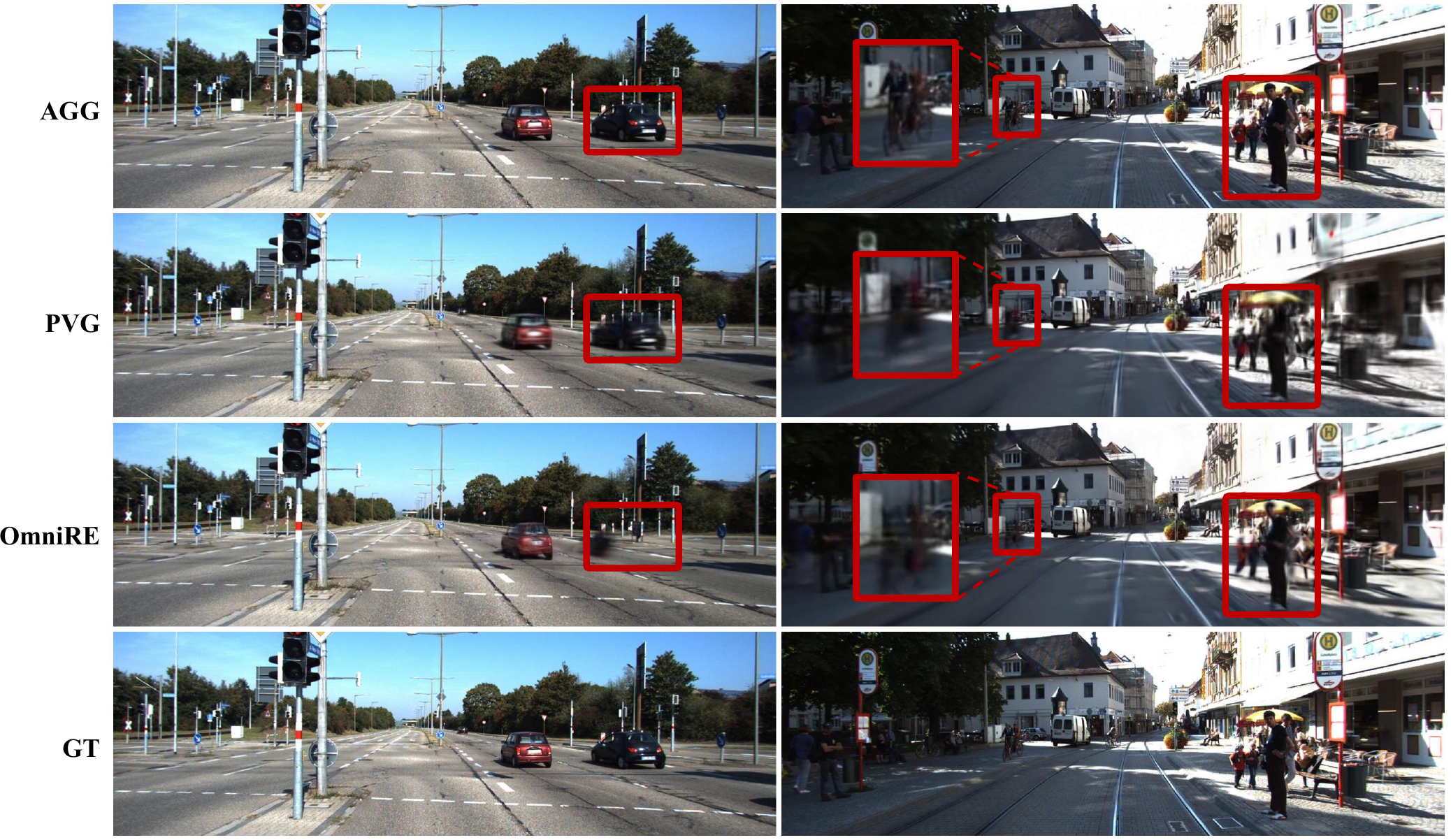}
\captionof{figure}{Qualitative comparisons on the KITTI dataset under noisy priors. \textbf{Please refer to the appendix for more visualization.}}
\label{fig:kitti-vis}
\end{table}
\noindent\textbf{Performance on KITTI.} 
Tab.~\ref{tab:kitti-metric} reports results under two initialization conditions. Under the ideal setting with ground-truth inputs, AGG establishes itself as a high-fidelity 4D reconstruction method, achieving state-of-the-art scene reconstruction and highly competitive NVS. While our NVS PSNR is slightly lower than EMD by 0.06, AGG outperforms it across all other metrics. This highlights a key advantage: since even high-quality human annotations contain minor errors, such as slight bounding box shifts, AGG actively corrects these flaws through stable optimization rather than overfitting.

Under the noisy setting with estimated priors, fixed-topology GSG methods including StreetGS and OmniRe, degrade severely, as they cannot recover from noisy priors. In contrast, AGG maintains robust performance, outperforming PVG by a large margin of 1.94 PSNR. Furthermore, Tab.~\ref{tab:kitti-agent} details agent-level metrics under this noisy setting, where AGG achieves 28.68 PSNR on vehicles and 33.57 PSNR on humans. These quantitative gains are strongly supported by our visualization in Fig.~\ref{fig:kitti-vis}. While PVG and OmniRe exhibit severe ghosting and blurring on dynamic objects (e.g., the black car and the pedestrians), AGG reconstructs them with fine details.
These results validate our core designs: Adaptive Topology Evolution corrects flawed initializations, and the Tick-Tock strategy resolves optimization ambiguities. This robustness enables AGG to scale to in-the-wild internet and generative videos. Regarding efficiency, AGG renders at 35 FPS, easily exceeding the 10 Hz real-time requirement for autonomous driving simulation despite being slower than vanilla 3DGS.

\noindent\textbf{Performance on Wild-30.} 
To explicitly validate this performance on highly unconstrained internet and generative videos, we evaluate the approach on the Wild-30 benchmark. As shown in Tab.~\ref{tab:wild30-metric}, standard baselines struggle to converge, leading to background artifacts and inaccurate agents. Conversely, AGG achieves a superior 30.14 full-image PSNR and significantly outperforms other methods across all dynamic-region metrics. Qualitatively, as illustrated in Fig.~\ref{fig:wild-vis}, baselines severely entangle ego-motion with dynamic objects. AGG successfully disentangles the scene, firmly demonstrating its capability for robust 4D reconstruction directly from in-the-wild driving videos. Furthermore, we provide extended comparisons with additional methods, such as recent feed-forward approaches, on the Wild-30 benchmark in the Appendix.

\begin{table}[!t]
\centering
\caption{Results on Wild-30 for scene reconstruction and novel view synthesis. \textbf{Bold} and \ul{underline} denote the best and second-best.}
\label{tab:wild30-metric}
\setlength{\tabcolsep}{3.8pt} 
\renewcommand{\arraystretch}{1.05} 
\resizebox{\textwidth}{!}{
\begin{tabular}{l ccc cc cc ccc cc cc}
\toprule

\multirow{3}{*}{Method} & \multicolumn{7}{c}{Scene Reconstruction} & \multicolumn{7}{c}{Novel View Synthesis (NVS)} \\ 
\cmidrule(lr){2-8} \cmidrule(lr){9-15} 

 & \multicolumn{3}{c}{Full} & \multicolumn{2}{c}{Vehicle} & \multicolumn{2}{c}{Human} & \multicolumn{3}{c}{Full} & \multicolumn{2}{c}{Vehicle} & \multicolumn{2}{c}{Human} \\ 
\cmidrule(lr){2-4} \cmidrule(lr){5-6} \cmidrule(lr){7-8} \cmidrule(lr){9-11} \cmidrule(lr){12-13} \cmidrule(lr){14-15} 

 & PSNR $\uparrow$ & SSIM $\uparrow$ & LPIPS $\downarrow$ & PSNR $\uparrow$ & SSIM $\uparrow$ & PSNR $\uparrow$ & SSIM $\uparrow$ & PSNR $\uparrow$ & SSIM $\uparrow$ & LPIPS $\downarrow$ & PSNR $\uparrow$ & SSIM $\uparrow$ & PSNR $\uparrow$ & SSIM $\uparrow$ \\ 
\midrule

DeformGS~\cite{deformgs}        & \ul{28.56} & \ul{0.899} & 0.107 & 23.62 & 0.764 & \ul{25.43} & \ul{0.793} & \ul{25.23} & \ul{0.802} & 0.145 & \ul{21.07} & 0.626 & \ul{22.36} & \ul{0.624} \\
PVG~\cite{chen2026pvg}          & 27.36 & 0.860 & 0.161 & 23.91 & 0.777 & 23.80 & 0.758 & 23.86 & 0.720 & 0.202 & 20.17 & 0.500 & 20.52 & 0.466 \\
OmniRe~\cite{chen2025omnire}    & 28.04 & 0.895 & \ul{0.106} & \ul{23.93} & \ul{0.783} & 25.16 & 0.789 & 24.74 & 0.796 & 0.142 & 21.05 & \ul{0.631} & 22.02 & 0.617 \\
StreetGS~\cite{yan2024streetgs} & 27.98 & 0.895 & 0.107 & 23.83 & 0.782 & 25.15 & 0.791 & 24.78 & 0.796 & \ul{0.141} & 20.89 & 0.628 & 22.07 & 0.618 \\

\midrule
\rowcolor{gray!15} 
\textbf{AGG (Ours)} & \textbf{30.14} & \textbf{0.913} & \textbf{0.090} & \textbf{26.09} & \textbf{0.833} & \textbf{27.42} & \textbf{0.818} & \textbf{26.22} & \textbf{0.831} & \textbf{0.119} & \textbf{22.02} & \textbf{0.671} & \textbf{23.50} & \textbf{0.643} \\ 

\bottomrule
\end{tabular}
}
\includegraphics[width=0.98\textwidth]{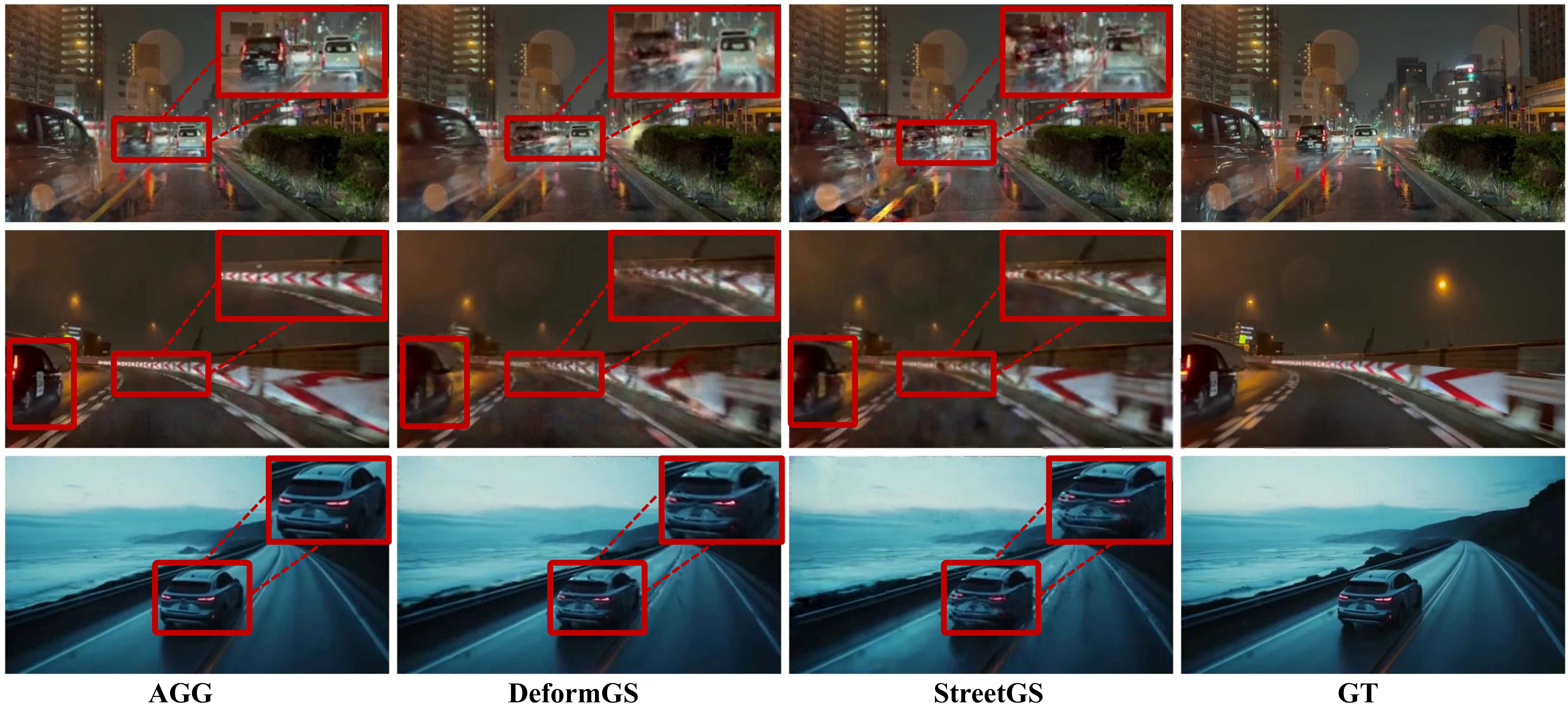}
\captionof{figure}{Qualitative comparisons on the Wild-30 dataset under noisy priors. \textbf{Please refer to the appendix for more visualization.}}
\label{fig:wild-vis}
\end{table}
\subsection{Robustness against noisy input}
\begin{figure}[!h]
    \centering
    \includegraphics[width=0.8\linewidth]{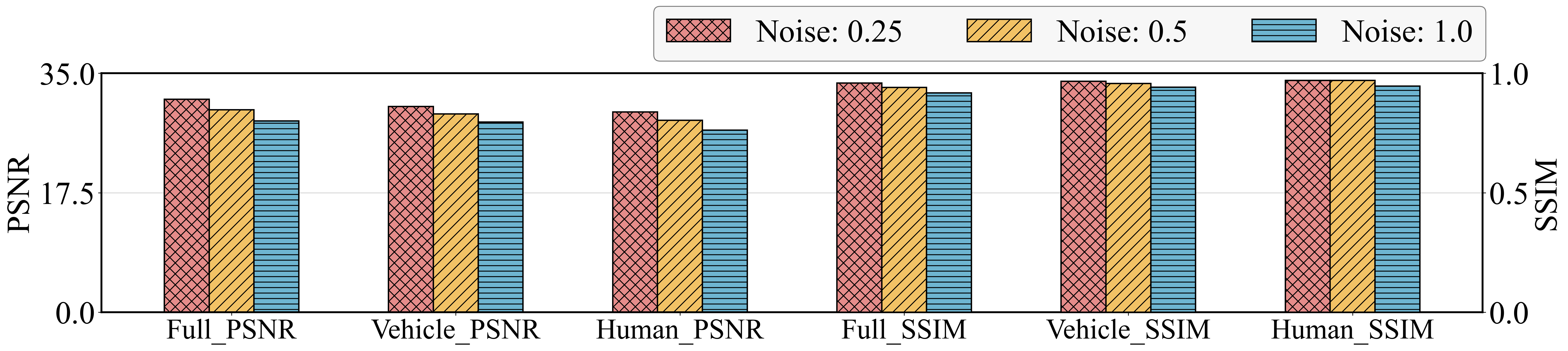}
    \caption{\textbf{Robustness against pose noise.} 
    % We report the reconstruction quality on a KITTI subset after injecting random translational noise (with maximum bounds of 0.25, 0.5, and 1.0 meters) into the initial camera poses. 
    AGG exhibits minimal performance drop even under severe 1.0-meter perturbations, highlighting the strong self-correction capability of our overall framework.}
    \label{fig:pose_noise}
\end{figure}

To further validate the robustness of AGG under noisy inputs, we conduct a stress test on a KITTI subset. We inject random translational noise into the accurate camera poses, specifically evaluating the reconstruction quality across three maximum noise bounds: 0.25, 0.5, and 1.0 meters. As shown in Fig.~\ref{fig:pose_noise}, AGG maintains highly stable performance. Even under a severe perturbation of 1.0 meter, the rendering metrics (PSNR and SSIM) for both the full image and dynamic agents exhibit only slight degradation. This degradation effectively proves that our decoupled optimization can successfully self-correct severe initialization errors without corrupting the scene geometry. Additional evaluations of AGG’s robustness under more diverse noise settings are provided in the Appendix.

\subsection{Ablation Study}

\begin{figure}[!h]
    \centering
    \begin{subfigure}[b]{0.48\linewidth}
        \centering
        % 请将 example-image-a 替换为你的真实图片路径，如 figures/method_part1.pdf
        \includegraphics[width=\linewidth]{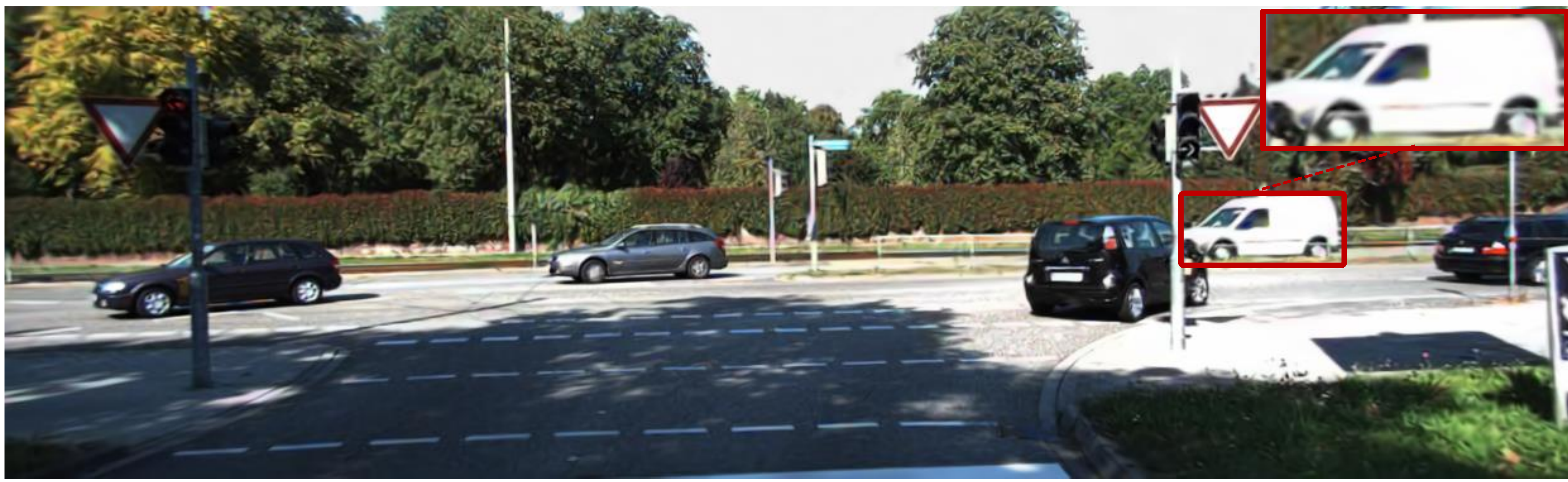} 
        \caption{Full model}
        \label{fig:abl-full}
    \end{subfigure}
    \hfill 
    \begin{subfigure}[b]{0.48\linewidth}
        \centering
        \includegraphics[width=\linewidth]{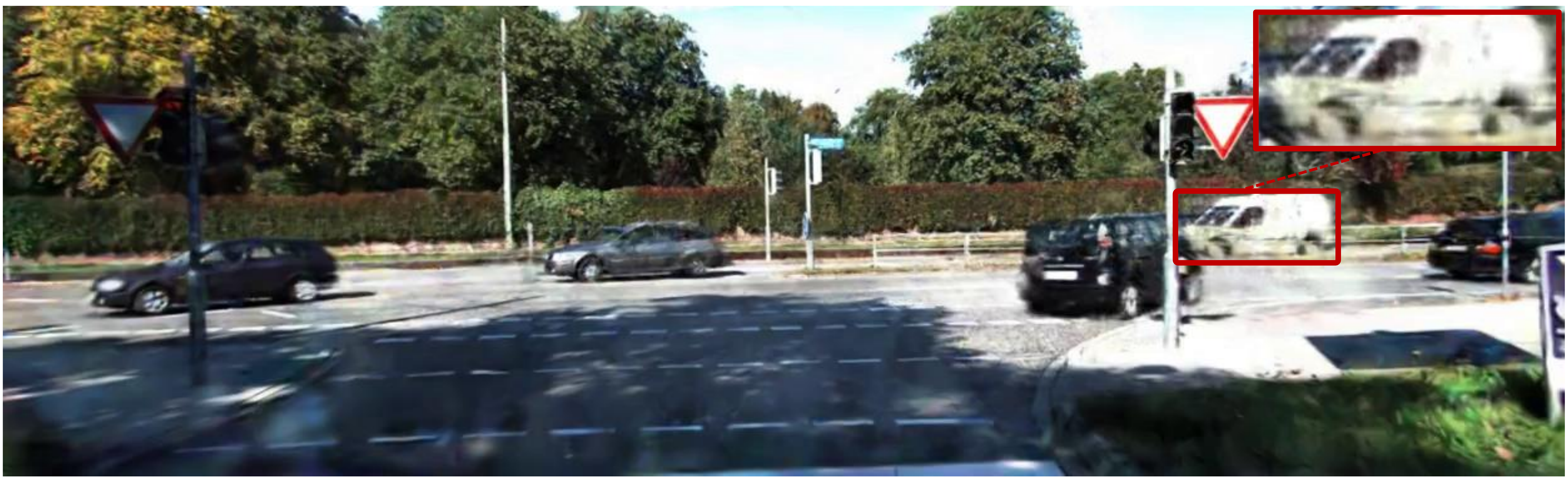} 
        \caption{w/o tick-tock strategy}
        \label{fig:abl-tick}
    \end{subfigure}
    \begin{subfigure}[b]{0.48\linewidth}
        \centering
        \includegraphics[width=\linewidth]{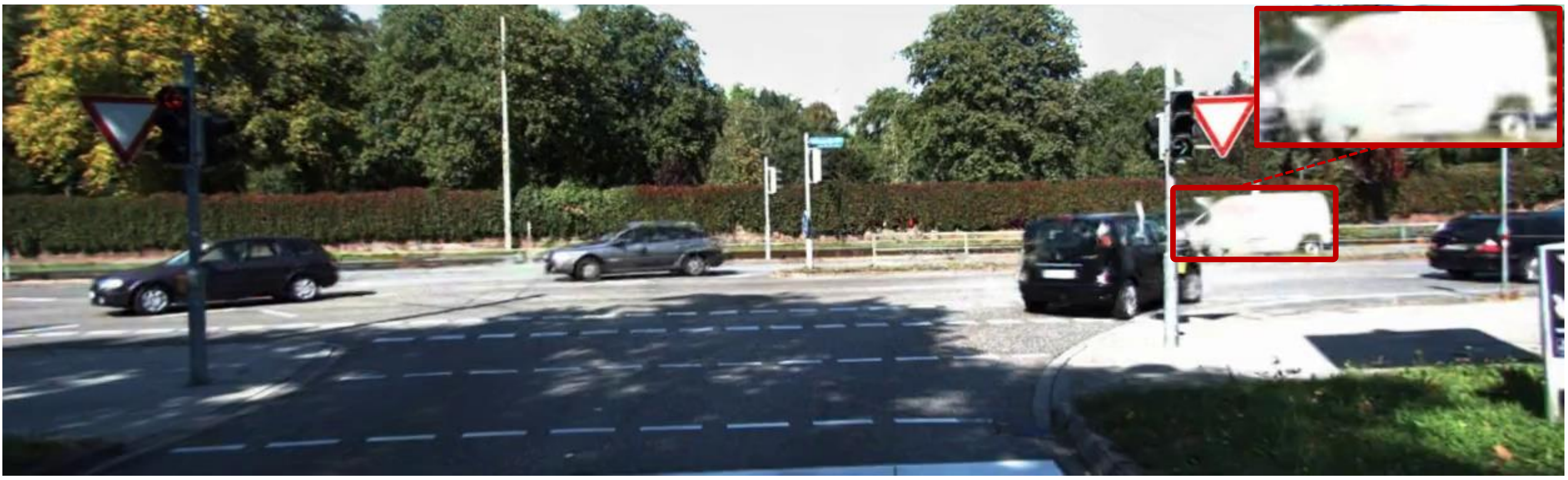}
        \caption{w/o topology evolution}
        \label{fig:abl-topo}
    \end{subfigure}
    \hfill
    \begin{subfigure}[b]{0.48\linewidth}
        \centering
        \includegraphics[width=\linewidth]{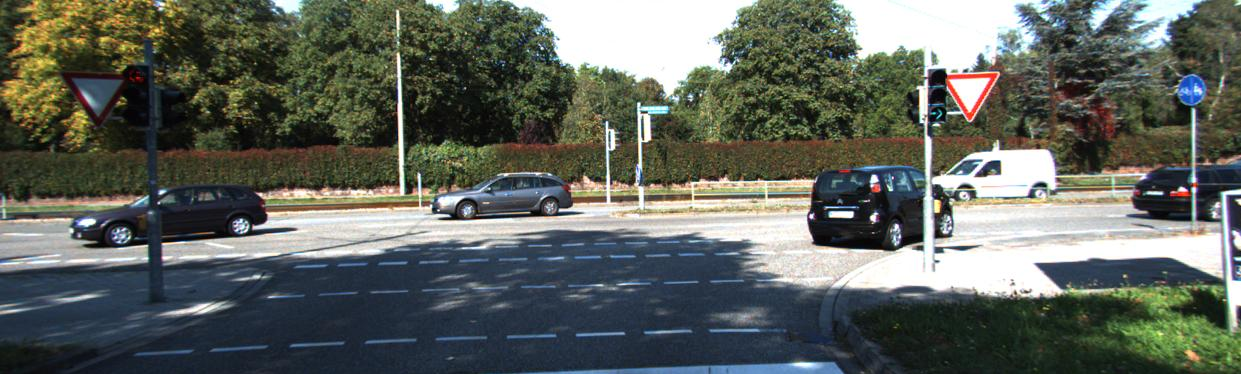}
        \caption{GT}
        \label{fig:abl-gt}
    \end{subfigure}
    \caption{Qualitative ablation study results from the KITTI dataset}
    \label{fig:abl}
\end{figure}

We ablate AGG’s core components on the KITTI dataset under noisy priors, with quantitative and qualitative results shown in  Tab.~\ref{tab:ablation_kitti} and Fig.~\ref{fig:abl}.
% We ablate the core components of AGG on the KITTI dataset under the noisy initialization setting. Quantitative and qualitative results are presented in Tab.~\ref{tab:ablation_kitti} and Fig.~\ref{fig:abl}, respectively.

\noindent\textbf{Tick-Tock Strategy.} 
Disabling the Tick-Tock strategy causes a significant performance drop of 1.92 PSNR (27.16 vs. 29.08). Jointly optimizing camera and agent poses creates severe ambiguity, blurring dynamic objects (Fig.~\ref{fig:abl-tick}). Our phase separation successfully decouples this optimization.

\noindent\textbf{Semantic Attention.} Removing this module drops PSNR to 27.15. Without semantic gradient routing, decoupled nodes lack clear optimization directions. As Fig.~\ref{fig:sem_att} shows, our attention maps ($C_{bg}$ and $C_{ag}$) stably highlight background and foreground regions to guide parameter updates.

\begin{table}[t] 
\centering
\setlength{\abovecaptionskip}{2pt} 
\setlength{\belowcaptionskip}{0pt} 
\begin{minipage}[t]{0.48\linewidth}
    \centering
    \caption{Agent Recon on KITTI.}
    \label{tab:kitti-agent}
    \resizebox{\linewidth}{!}{
    \begin{tabular}{lcccc}
    \hline
    \multirow{2}{*}{Method} & \multicolumn{2}{c}{Vehicle} & \multicolumn{2}{c}{Human} \\ 
    \cline{2-3} \cline{4-5} 
     & PSNR $\uparrow$ & SSIM $\uparrow$ & PSNR $\uparrow$ & SSIM $\uparrow$ \\ \hline
    PVG~\cite{chen2026pvg} & 26.15 & 0.818 & 28.55 & 0.867 \\
    DeformGS~\cite{deformgs} & 23.85 & 0.713 & 23.19 & 0.695 \\
    StreetGS~\cite{yan2024streetgs} & 25.85 & 0.805 & 30.97 & 0.922 \\
    OmniRe~\cite{chen2025omnire} & 25.48 & 0.824 & 27.50 & 0.962 \\
    \rowcolor{grayrow} \textbf{AGG (Ours)} & \textbf{28.68} & \textbf{0.868} & \textbf{33.57} & \textbf{0.988} \\ \hline
    \end{tabular}
    }
\end{minipage}%
\hfill
\begin{minipage}[t]{0.48\linewidth}
    \centering
    \caption{Ablation Study on KITTI.}
    \label{tab:ablation_kitti}
    \resizebox{\linewidth}{!}{
    \begin{tabular}{lccc}
    \hline
    \multirow{2}{*}{Method} & \multicolumn{3}{c}{Reconstruction} \\ 
    \cline{2-4}
     & PSNR $\uparrow$ & SSIM $\uparrow$ & LPIPS $\downarrow$ \\ \hline
    w/o tick-tock  & 27.16 & 0.911 &  0.069  \\ 
    w/o sem attention  & 27.15 & 0.911 &  0.070  \\
    w/o topology update  & 27.03 & 0.907 & 0.071 \\ 
    \rowcolor{grayrow} Full Model & 29.08 & 0.943 & 0.040   \\
    \hline
    \end{tabular}
    }
\end{minipage}
\end{table}

\begin{figure}[!h] 
\setlength{\abovecaptionskip}{2pt} 
\setlength{\belowcaptionskip}{0pt} 
    \centering
    \begin{subfigure}[b]{0.32\linewidth}
        \centering
        \includegraphics[width=\linewidth]{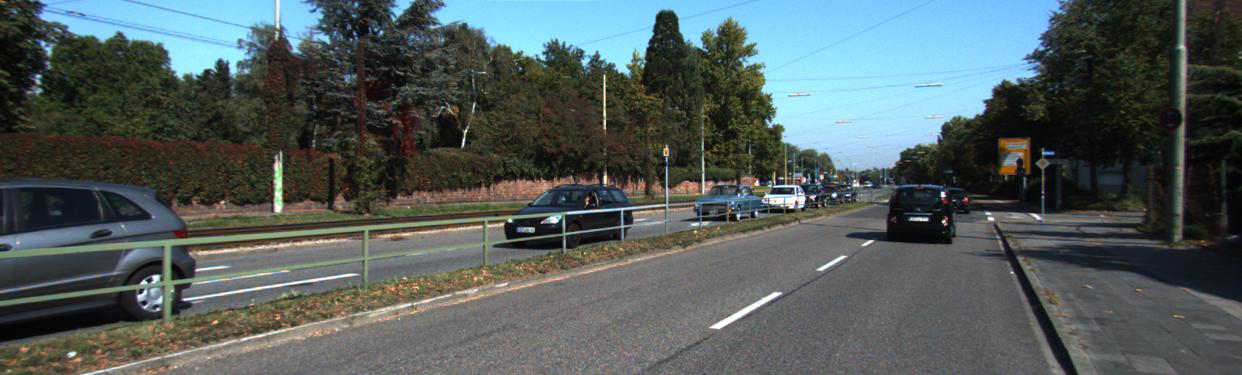}
        \caption{Input}
    \end{subfigure}
    \hfill 
    \begin{subfigure}[b]{0.32\linewidth}
        \centering
        \includegraphics[width=\linewidth]{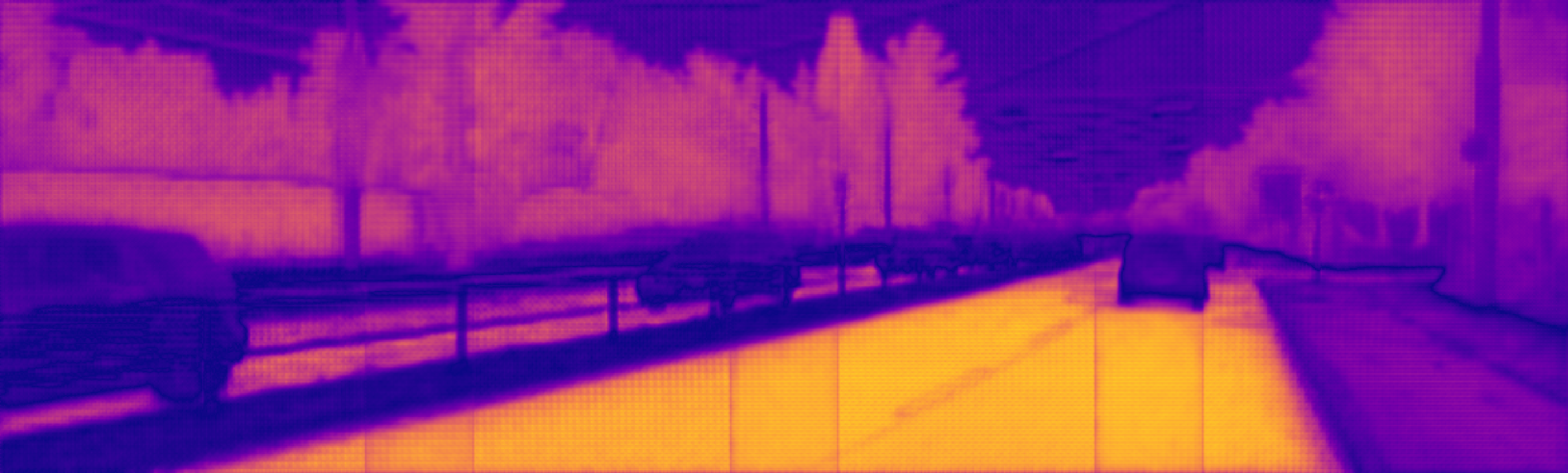}
        \caption{Static Attention ($C_{bg}$)}
    \end{subfigure}
    \hfill 
    \begin{subfigure}[b]{0.32\linewidth}
        \centering
        \includegraphics[width=\linewidth]{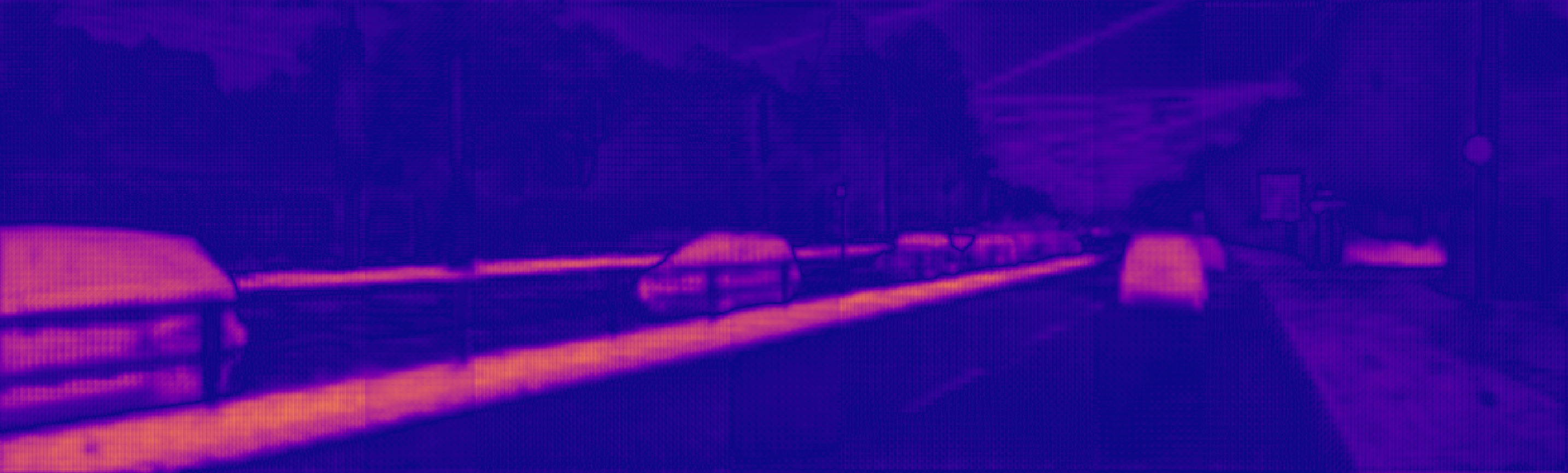}
        \caption{Dynamic Attention ($C_{ag}$)}
    \end{subfigure}
    \caption{Visualization of semantic attention maps.}
    \label{fig:sem_att}
\end{figure}

\noindent\textbf{Adaptive Topology Evolution.} 
Removing the Adaptive Topology Evolution (ATE) causes the largest PSNR drop of 2.05.  Bound to a flawed initial topology, the model cannot recover missing agents. For instance, a missed white van forces background Gaussians to over-stretch into a blurry blob (Fig.~\ref{fig:abl-topo}). ATE actively corrects these initialization failures by spawning new agent nodes. Prototype bank evaluations are in the Appendix.

\section{Conclusion}
\label{sec:conclusion}
In this paper, we introduce Adaptive Gaussian Graph (AGG), a zero-cost, self-correcting 4D reconstruction framework designed to overcome the diversity bottleneck of existing Gaussian Scene Graph methods. Rather than passively inheriting errors from noisy, off-the-shelf priors, AGG actively rectifies them. It achieves this through a Semantically Guided Tick-Tock Strategy that decouples static and dynamic optimization, and an Adaptive Topology Evolution module that fixes topology flaws. We also present Wild-30, a new benchmark featuring internet and generative videos. Comprehensive experiments demonstrate that AGG significantly outperforms state-of-the-art methods, unlocking robust in-the-wild scene reconstruction from noisy priors for diverse driving simulations. 

\newpage
\bibliographystyle{splncs04}
\bibliography{main}

@String(CVPR  = {IEEE Conf. Comput. Vis. Pattern Recog.})

@String(ECCV  = {Eur. Conf. Comput. Vis.})

@String(ICLR  = {Int. Conf. Learn. Represent.})

@String(CVPR  = {CVPR})

@String(ECCV  = {ECCV})

@String(ICLR  = {ICLR})

@Article{kerbl3Dgaussians,
      author       = {Kerbl, Bernhard and Kopanas, Georgios and Leimk{\"u}hler, Thomas and Drettakis, George},
      title        = {3D Gaussian Splatting for Real-Time Radiance Field Rendering},
      journal      = {ACM Transactions on Graphics},
      number       = {4},
      volume       = {42},
      month        = {July},
      year         = {2023},
      url          = {https://repo-sam.inria.fr/fungraph/3d-gaussian-splatting/}
}

@inproceedings{yan2024streetgs,
    title={Street Gaussians: Modeling Dynamic Urban Scenes with Gaussian Splatting}, 
    author={Yunzhi Yan and Haotong Lin and Chenxu Zhou and Weijie Wang and Haiyang Sun and Kun Zhan and Xianpeng Lang and Xiaowei Zhou and Sida Peng},
    booktitle={ECCV},
    year={2024}
}

@inproceedings{
  chen2025omnire,
  title={OmniRe: Omni Urban Scene Reconstruction},
  author={Ziyu Chen and Jiawei Yang and Jiahui Huang and Riccardo de Lutio and Janick Martinez Esturo and Boris Ivanovic and Or Litany and Zan Gojcic and Sanja Fidler and Marco Pavone and Li Song and Yue Wang},
  booktitle={The Thirteenth International Conference on Learning Representations},
  year={2025}
}

@inproceedings{zhou2024drivinggaussian,
  title={Drivinggaussian: Composite gaussian splatting for surrounding dynamic autonomous driving scenes},
  author={Zhou, Xiaoyu and Lin, Zhiwei and Shan, Xiaojun and Wang, Yongtao and Sun, Deqing and Yang, Ming-Hsuan},
  booktitle={Proceedings of the IEEE/CVF Conference on Computer Vision and Pattern Recognition},
  pages={21634--21643},
  year={2024}
}

@inproceedings{huang2025vipe,
    title={ViPE: Video Pose Engine for 3D Geometric Perception},
    author={Huang, Jiahui and Zhou, Qunjie and Rabeti, Hesam and Korovko, Aleksandr and Ling, Huan and Ren, Xuanchi and Shen, Tianchang and Gao, Jun and Slepichev, Dmitry and Lin, Chen-Hsuan and Ren, Jiawei and Xie, Kevin and Biswas, Joydeep and Leal-Taixe, Laura and Fidler, Sanja},
    booktitle={NVIDIA Research Whitepapers arXiv:2508.10934},
    year={2025}
}

@inproceedings{xie2023snerf,
author = {Xie, Ziyang and Zhang, Junge and Li, Wenye and Zhang, Feihu and Zhang, Li},
title = {S-NeRF: Neural Radiance Fields for Street Views},
booktitle = {International Conference on Learning Representations (ICLR)},
year = {2023}
}

@article{guo2023streetsurf,
    title={StreetSurf: Extending Multi-view Implicit Surface Reconstruction to Street Views},
    author={Guo, Jianfei and Deng, Nianchen and Li, Xinyang and Bai, Yeqi and Shi, Botian and Wang, Chiyu and Ding, Chenjing and Wang, Dongliang and Li, Yikang},
    journal={arXiv preprint arXiv:2306.04988},
    year={2023}
}

@InProceedings{Ost_2021_NSG,
    author    = {Ost, Julian and Mannan, Fahim and Thuerey, Nils and Knodt, Julian and Heide, Felix},
    title     = {Neural Scene Graphs for Dynamic Scenes},
    booktitle = {Proceedings of the IEEE/CVF Conference on Computer Vision and Pattern Recognition (CVPR)},
    month     = {June},
    year      = {2021},
    pages     = {2856-2865}
}

@inproceedings{mars,
  title={Space-time neural irradiance fields for free-viewpoint video},
  author={Xian, Wenqi and Huang, Jia-Bin and Kopf, Johannes and Kim, Changil},
  booktitle={Proceedings of the IEEE/CVF conference on computer vision and pattern recognition},
  pages={9421--9431},
  year={2021}
}

@inproceedings{turki2023suds,
  title={Suds: Scalable urban dynamic scenes},
  author={Turki, Haithem and Zhang, Jason Y and Ferroni, Francesco and Ramanan, Deva},
  booktitle={Proceedings of the IEEE/CVF Conference on Computer Vision and Pattern Recognition},
  pages={12375--12385},
  year={2023}
}

@article{yang2023emernerf,
  title={Emernerf: Emergent spatial-temporal scene decomposition via self-supervision},
  author={Yang, Jiawei and Ivanovic, Boris and Litany, Or and Weng, Xinshuo and Kim, Seung Wook and Li, Boyi and Che, Tong and Xu, Danfei and Fidler, Sanja and Pavone, Marco and others},
  journal={arXiv preprint arXiv:2311.02077},
  year={2023}
}

@article{chen2026pvg,
  title={Periodic vibration gaussian: Dynamic urban scene reconstruction and real-time rendering},
  author={Chen, Yurui and Gu, Chun and Jiang, Junzhe and Zhu, Xiatian and Zhang, Li},
  journal={International Journal of Computer Vision},
  volume={134},
  number={3},
  pages={83},
  year={2026},
  publisher={Springer}
}

@inproceedings{peng2025desire,
  title={Desire-gs: 4d street gaussians for static-dynamic decomposition and surface reconstruction for urban driving scenes},
  author={Peng, Chensheng and Zhang, Chengwei and Wang, Yixiao and Xu, Chenfeng and Xie, Yichen and Zheng, Wenzhao and Keutzer, Kurt and Tomizuka, Masayoshi and Zhan, Wei},
  booktitle={Proceedings of the Computer Vision and Pattern Recognition Conference},
  pages={6782--6791},
  year={2025}
}

@inproceedings{wei2025emd,
  title={Emd: Explicit motion modeling for high-quality street gaussian splatting},
  author={Wei, Xiaobao and Wuwu, Qingpo and Zhao, Zhongyu and Wu, Zhuangzhe and Huang, Nan and Lu, Ming and Ma, Ningning and Zhang, Shanghang},
  booktitle={Proceedings of the IEEE/CVF international conference on computer vision},
  pages={28462--28472},
  year={2025}
}

@inproceedings{Cordts2016Cityscapes,
title={The Cityscapes Dataset for Semantic Urban Scene Understanding},
author={Cordts, Marius and Omran, Mohamed and Ramos, Sebastian and Rehfeld, Timo and Enzweiler, Markus and Benenson, Rodrigo and Franke, Uwe and Roth, Stefan and Schiele, Bernt},
booktitle={Proc. of the IEEE Conference on Computer Vision and Pattern Recognition (CVPR)},
year={2016}
}

@inproceedings{tancik2022blocknerf,
  title={Block-nerf: Scalable large scene neural view synthesis},
  author={Tancik, Matthew and Casser, Vincent and Yan, Xinchen and Pradhan, Sabeek and Mildenhall, Ben and Srinivasan, Pratul P and Barron, Jonathan T and Kretzschmar, Henrik},
  booktitle={Proceedings of the IEEE/CVF conference on computer vision and pattern recognition},
  pages={8248--8258},
  year={2022}
}

@article{zhou2024hugsim,
  title={HUGSIM: A Real-Time, Photo-Realistic and Closed-Loop Simulator for Autonomous Driving},
  author={Zhou, Hongyu and Lin, Longzhong and Wang, Jiabao and Lu, Yichong and Bai, Dongfeng and Liu, Bingbing and Wang, Yue and Geiger, Andreas and Liao, Yiyi},
  journal={arXiv preprint arXiv:2412.01718},
  year={2024}
}

@inproceedings{hu2024-DepthCrafter,
author      = {Hu, Wenbo and Gao, Xiangjun and Li, Xiaoyu and Zhao, Sijie and Cun, Xiaodong and Zhang, Yong and Quan, Long and Shan, Ying},
title       = {DepthCrafter: Generating Consistent Long Depth Sequences for Open-world Videos},
booktitle   = {CVPR},
year        = {2025}
}

@inproceedings{must3r_cvpr25,
      title={MUSt3R: Multi-view Network for Stereo 3D Reconstruction}, 
      author={Yohann Cabon and Lucas Stoffl and Leonid Antsfeld and Gabriela Csurka and Boris Chidlovskii and Jerome Revaud and Vincent Leroy},
      booktitle = {CVPR},
      year = {2025}
}

@InProceedings{Zhou2024HUGS,
    author    = {Zhou, Hongyu and Shao, Jiahao and Xu, Lu and Bai, Dongfeng and Qiu, Weichao and Liu, Bingbing and Wang, Yue and Geiger, Andreas and Liao, Yiyi},
    title     = {HUGS: Holistic Urban 3D Scene Understanding via Gaussian Splatting},
    booktitle = {Proceedings of the IEEE/CVF Conference on Computer Vision and Pattern Recognition (CVPR)},
    month     = {June},
    year      = {2024},
    pages     = {21336-21345}
}

@inproceedings{huang2026s3gaussian,
  title={S3Gaussian: Self-Supervised Street Gaussians for Autonomous Driving},
  author={Huang, Nan and Wei, Xiaobao and Zheng, Wenzhao and An, Pengju and Lu, Ming and Zhan, Wei and Tomizuka, Masayoshi and Keutzer, Kurt and Zhang, Shanghang},
  booktitle={Proceedings of the IEEE International Conference on Robotics and Automation (ICRA)},
  year={2026},
}

@InProceedings{waymo, 
author = {Sun, Pei and Kretzschmar, Henrik and Dotiwalla, Xerxes and Chouard, Aurelien and Patnaik, Vijaysai and Tsui, Paul and Guo, James and Zhou, Yin and Chai, Yuning and Caine, Benjamin and Vasudevan, Vijay and Han, Wei and Ngiam, Jiquan and Zhao, Hang and Timofeev, Aleksei and Ettinger, Scott and Krivokon, Maxim and Gao, Amy and Joshi, Aditya and Zhang, Yu and Shlens, Jonathon and Chen, Zhifeng and Anguelov, Dragomir}, 
title = {Scalability in Perception for Autonomous Driving: Waymo Open Dataset}, 
booktitle = {Proceedings of the IEEE/CVF Conference on Computer Vision and Pattern Recognition (CVPR)},
month = {June}, 
year = {2020}
}

@inproceedings{kitti,
  author = {Andreas Geiger and Philip Lenz and Raquel Urtasun},
  title = {Are we ready for Autonomous Driving? The KITTI Vision Benchmark Suite},
  booktitle = {Conference on Computer Vision and Pattern Recognition (CVPR)},
  year = {2012}
}

@inproceedings{wang2025vggt,
  title={VGGT: Visual Geometry Grounded Transformer},
  author={Wang, Jianyuan and Chen, Minghao and Karaev, Nikita and Vedaldi, Andrea and Rupprecht, Christian and Novotny, David},
  booktitle={Proceedings of the IEEE/CVF Conference on Computer Vision and Pattern Recognition},
  year={2025}
}

@misc{hu2025vggt4d,
      author={Yu Hu and Chong Cheng and Sicheng Yu and Xiaoyang Guo and Hao Wang},
      year={2025},
      eprint={2511.19971},
      archivePrefix={arXiv},
      primaryClass={cs.CV},
      url={https://arxiv.org/abs/2511.19971}, 
}

@article{SE3,
  title = {Dynamic Modeling and Locomotion Control for Quadruped Robots Based on Center of Inertia on {{SE}}(3)},
  author = {Ding, Xilun and Chen, Hao},
  date = {2015-10},
  journaltitle = {Journal of Dynamic Systems, Measurement, and Control},
  volume = {138},
  number = {1},
  eprint = {https://asmedigitalcollection.asme.org/dynamicsystems/article-pdf/138/1/011004/6120208/ds_138_01_011004.pdf},
  pages = {011004},
  issn = {0022-0434},
  doi = {10.1115/1.4031728},
  url = {https://doi.org/10.1115/1.4031728}
}

@misc{simeoni2025dinov3,
  title={{DINOv3}},
  author={Sim{\'e}oni, Oriane and Vo, Huy V. and Seitzer, Maximilian and Baldassarre, Federico and Oquab, Maxime and Jose, Cijo and Khalidov, Vasil and Szafraniec, Marc and Yi, Seungeun and Ramamonjisoa, Micha{\"e}l and Massa, Francisco and Haziza, Daniel and Wehrstedt, Luca and Wang, Jianyuan and Darcet, Timoth{\'e}e and Moutakanni, Th{\'e}o and Sentana, Leonel and Roberts, Claire and Vedaldi, Andrea and Tolan, Jamie and Brandt, John and Couprie, Camille and Mairal, Julien and J{\'e}gou, Herv{\'e} and Labatut, Patrick and Bojanowski, Piotr},
  year={2025},
  eprint={2508.10104},
  archivePrefix={arXiv},
  primaryClass={cs.CV},
  url={https://arxiv.org/abs/2508.10104},
}

@article{abdi2010pca,
  title={Principal component analysis},
  author={Abdi, Herv{\'e} and Williams, Lynne J},
  journal={Wiley interdisciplinary reviews: computational statistics},
  volume={2},
  number={4},
  pages={433--459},
  year={2010},
  publisher={Wiley Online Library}
}

@article{kf1960,
    Author = {Kalman, Rudolph Emil},
    Title = {A New Approach to Linear Filtering and Prediction Problems},
    Journal = {Transactions of the ASME--Journal of Basic Engineering},
    Volume = {82},
    Number = {Series D},
    Pages = {35--45},
    Year = {1960}
}

@ARTICLE{scs1,
  author={Ding, Wenhao and Xu, Chejian and Arief, Mansur and Lin, Haohong and Li, Bo and Zhao, Ding},
  journal={IEEE Transactions on Intelligent Transportation Systems}, 
  title={A Survey on Safety-Critical Driving Scenario Generation—A Methodological Perspective}, 
  year={2023},
  volume={24},
  number={7},
  pages={6971-6988},
  doi={10.1109/TITS.2023.3259322}}

@article{scs2,
  title = {Corner Cases in Machine Learning Processes},
  author = {Heidecker, Florian and Bieshaar, Maarten and Sick, Bernhard},
  date = {2024-01-02},
  journaltitle = {AI Perspectives \& Advances},
  shortjournal = {AI Perspect. Adv.},
  volume = {6},
  number = {1},
  pages = {1},
  issn = {2948-2143},
  doi = {10.1186/s42467-023-00015-y},
  url = {https://doi.org/10.1186/s42467-023-00015-y},
  urldate = {2025-03-05},
  langid = {english}
}

@inproceedings{scs3,
  title={Coda: A real-world road corner case dataset for object detection in autonomous driving},
  author={Li, Kaican and Chen, Kai and Wang, Haoyu and Hong, Lanqing and Ye, Chaoqiang and Han, Jianhua and Chen, Yukuai and Zhang, Wei and Xu, Chunjing and Yeung, Dit-Yan and others},
  booktitle={European Conference on Computer Vision},
  pages={406--423},
  year={2022},
  organization={Springer}
}

@inproceedings{unisim,
  title = {{{UniSim}}: {{A}} Neural Closed-Loop Sensor Simulator},
  shorttitle = {{{UniSim}}},
  booktitle = {2023 {{IEEE}}/{{CVF Conference}} on {{Computer Vision}} and {{Pattern Recognition}} ({{CVPR}})},
  author = {Yang, Ze and Chen, Yun and Wang, Jingkang and Manivasagam, Sivabalan and Ma, Wei-Chiu and Yang, Anqi Joyce and Urtasun, Raquel},
  date = {2023-06},
  pages = {1389--1399},
  publisher = {IEEE},
  location = {Vancouver, BC, Canada},
  doi = {10.1109/CVPR52729.2023.00140},
  url = {https://ieeexplore.ieee.org/document/10204923/},
  urldate = {2024-11-12},
  eventtitle = {2023 {{IEEE}}/{{CVF Conference}} on {{Computer Vision}} and {{Pattern Recognition}} ({{CVPR}})},
  isbn = {979-8-3503-0129-8},
  langid = {english},
}

@online{oasim,
  title = {{{OASim}}: {{An}} Open and Adaptive Simulator Based on Neural Rendering for Autonomous Driving},
  shorttitle = {{{OASim}}},
  author = {Yan, Guohang and Pi, Jiahao and Guo, Jianfei and Luo, Zhaotong and Dou, Min and Deng, Nianchen and Huang, Qiusheng and Fu, Daocheng and Wen, Licheng and Cai, Pinlong and Gao, Xing and Cai, Xinyu and Zhang, Bo and Yang, Xuemeng and Bai, Yeqi and Zhou, Hongbin and Shi, Botian},
  date = {2024-02-06},
  eprint = {2402.03830},
  eprinttype = {arXiv},
  doi = {10.48550/arXiv.2402.03830},
  url = {http://arxiv.org/abs/2402.03830},
  urldate = {2024-11-12},
  langid = {american},
  pubstate = {prepublished}
}

@inproceedings{chen2021geosim,
  title={Geosim: Realistic video simulation via geometry-aware composition for self-driving},
  author={Chen, Yun and Rong, Frieda and Duggal, Shivam and Wang, Shenlong and Yan, Xinchen and Manivasagam, Sivabalan and Xue, Shangjie and Yumer, Ersin and Urtasun, Raquel},
  booktitle={Proceedings of the IEEE/CVF conference on computer vision and pattern recognition},
  pages={7230--7240},
  year={2021}
}

@INPROCEEDINGS{deformgs,
  author={Yang, Ziyi and Gao, Xinyu and Zhou, Wen and Jiao, Shaohui and Zhang, Yuqing and Jin, Xiaogang},
  booktitle={2024 IEEE/CVF Conference on Computer Vision and Pattern Recognition (CVPR)}, 
  title={Deformable 3D Gaussians for High-Fidelity Monocular Dynamic Scene Reconstruction}, 
  year={2024},
  volume={},
  number={},
  pages={20331-20341},
  doi={10.1109/CVPR52733.2024.01922}}

@online{youtube,
  author = {YouTube},
  title = {YouTube},
  year = {2026},
  url = {https://www.youtube.com/}
}

@online{sora,
  author = {OpenAI},
  title = {Sora: Creating Video from Text},
  year = {2026},
  url = {https://openai.com/sora}
}

@inproceedings{kerssies2025eomt,
  author    = {Kerssies, Tommie and Cavagnero, Niccol\`{o} and Hermans, Alexander and Norouzi, Narges and Averta, Giuseppe and Leibe, Bastian and Dubbelman, Gijs and {de Geus}, Daan},
  title     = {{Your ViT is Secretly an Image Segmentation Model}},
  booktitle = {Proceedings of the IEEE/CVF Conference on Computer Vision and Pattern Recognition (CVPR)},
  year      = {2025},
}

@inproceedings{lpips,
  title={The unreasonable effectiveness of deep features as a perceptual metric},
  author={Zhang, Richard and Isola, Phillip and Efros, Alexei A and Shechtman, Eli and Wang, Oliver},
  booktitle={Proceedings of the IEEE conference on computer vision and pattern recognition},
  pages={586--595},
  year={2018}
}

@article{ssim,
  title={Image quality assessment: from error visibility to structural similarity},
  author={Wang, Zhou and Bovik, Alan C and Sheikh, Hamid R and Simoncelli, Eero P},
  journal={IEEE transactions on image processing},
  volume={13},
  number={4},
  pages={600--612},
  year={2004},
  publisher={IEEE}
}
\end{document}